\documentclass[journal]{IEEEtran}

\usepackage[utf8]{inputenc}
\usepackage{siunitx}
\usepackage[colorlinks=true, allcolors=blue]{hyperref}
\usepackage{amsmath}
\usepackage{amsfonts}
\usepackage{soul} 
\usepackage{subcaption}
\usepackage{stfloats}
\usepackage{booktabs}
\usepackage{enumitem}
\usepackage{colortbl}

\usepackage{tikz,xcolor}

\definecolor{lime}{HTML}{A6CE39}
\DeclareRobustCommand{\orcidicon}
{
    \begin{tikzpicture}
    \draw[lime, fill=lime] (0,0) circle [radius=0.16] 
    node[white] {{\fontfamily{qag}\selectfont \tiny ID}};    \draw[white, fill=white] (-0.0625,0.095) circle [radius=0.007];    
    \end{tikzpicture}
    \hspace{0mm}}
\foreach \x in {A, ..., Z}{%
    \expandafter\xdef\csname orcid\x\endcsname{\noexpand\href{https://orcid.org/\csname orcidauthor\x\endcsname}{\noexpand\orcidicon}}
}

\setlist[enumerate]{itemsep = 0pt, parsep = 0pt, topsep = 0pt} 
\setlist[itemize]{itemsep = 0pt, parsep = 0pt, topsep = 0pt} 

\begin{document}

\title{Deep Reinforcement Learning for Adverse Garage Scenario Generation}

\author{Kai Li \vspace{-20pt}}

\markboth{Journal of \LaTeX\ Class Files,~Vol.
}%
{Shell \MakeLowercase{\textit{et al.}}: Bare Demo of IEEEtran.cls for IEEE Journals}

\maketitle
\begin{abstract}
Autonomous vehicles need to travel over 11 billion miles to ensure their safety. Therefore, the importance of simulation testing before real-world testing is self-evident. 
In recent years, the release of 3D simulators for autonomous driving, represented by Carla and CarSim, marks the transition of autonomous driving simulation testing environments from simple 2D overhead views to complex 3D models.
During simulation testing, experimenters need to build static scenes and dynamic traffic flows, pedestrian flows, and other experimental elements to construct experimental scenarios. 
When building static scenes in 3D simulators, experimenters often need to manually construct 3D models, set parameters and attributes, which is time-consuming and labor-intensive. 
This thesis proposes an automated program generation framework. Based on deep reinforcement learning, this framework can generate different 2D ground script codes, on which 3D model files and map model files are built.
The generated 3D ground scenes are displayed in the Carla simulator, where experimenters can use this scene for navigation algorithm simulation testing.

\end{abstract}


%
\IEEEpeerreviewmaketitle

\section{Introduction}
\subsection{Background}
\begin{figure}[!ht]  \centering
    \includegraphics[width=.45\textwidth,trim={150 20 180 20},clip]{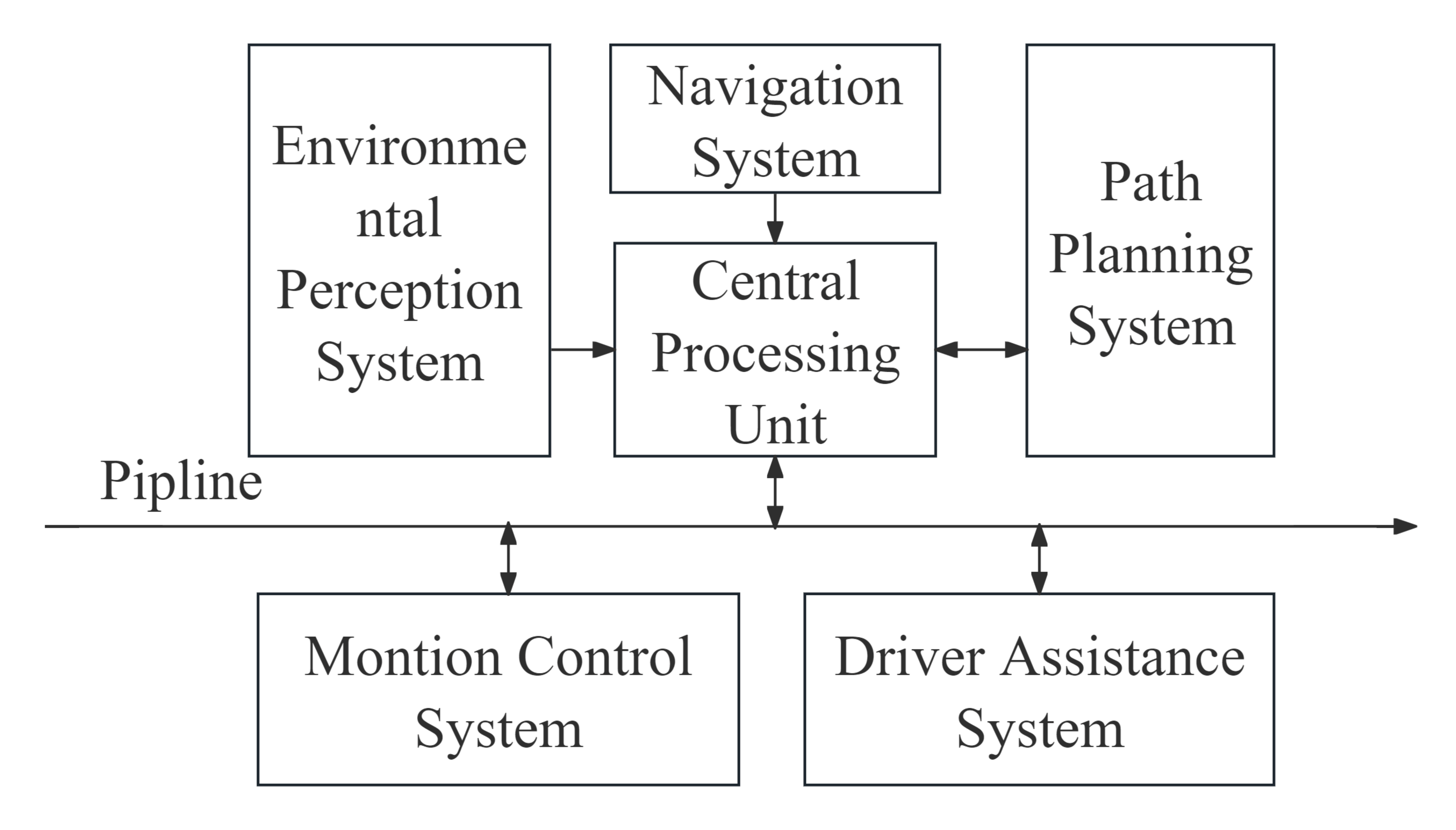}
    \caption{The architecture of an autonomous driving system.}\label{F:ads-architecture}
\end{figure}

The Self-Driving System, also known as the Autonomous Driving System (ADS), is a comprehensive integration of hardware and software designed to autonomously manage motion control based on its perception and understanding of surrounding environmental conditions, aiming to achieve a level of competence comparable to that of a human driver.
Perception, decision-making, and control constitute the three fundamental components of an autonomous driving system. As illustrated in \autoref{F:ads-architecture}, the architecture of an autonomous driving system encompasses the environmental perception system, the positioning and navigation system, the path planning system, the motion control system, and the driver assistance system \cite{lan2023end}.
The environmental perception system processes data collected from various sensors and transmits it to the central processing unit (CPU). 
A navigation system, such as GPS, is used to determine the vehicle's location within its environment. 
The path planning system is responsible for determining the vehicle's travel path, typically utilizing electronic maps. After integrating all the data, the CPU sends control commands to the motion control system, which manages specific vehicle movements, such as acceleration, braking, and steering.
Driver assistance system comprises a range of subsystems that aid in completing driving tasks, including automatic parking systems and emergency braking systems.

In ADS, the performance of autonomous driving relies on data collected from various types of sensors. 
This data is processed by algorithms, which ultimately determine the behavior of the autonomous driving system. 
Consequently, a diverse set of sensor data can enhance the performance of the autonomous driving system.

However, experiments have shown that autonomous vehicles need to travel over 11 billion miles to ensure their safety \cite{kalra2016driving}. 
In practical use and testing, traffic accidents caused by autonomous vehicles often place the ADS under intense scrutiny. As one of the most critical quality assurance technologies, ADS testing has garnered attention from both academia and industry \cite{test-begin}. 
Nonetheless, due to the numerous components and high complexity of ADS, testing faces many challenges.

Currently, ADS testing methods are gradually shifting from Naturalistic Field Operational Testing (N-FOT) to simulation-based testing \cite{khastgir2015identifying}, also known as simulation testing.
Simulation testing involves simulating various vehicle and environmental information within a computer system and verifying whether the autonomous driving algorithms operate as intended. 
During simulation testing, the autonomous driving system makes decisions based on sensor data collected in the simulated environment, and these decisions are reflected in the behavior of the vehicle controlled within the simulation, thereby achieving an effect similar to real vehicle testing. 
Compared to real vehicle testing, simulation testing consumes fewer resources and is more efficient in training, making it particularly useful for unit testing of algorithms, which targets specific functions.

Autonomous driving simulation software provides the necessary testing environment for simulation-based tests. 
In recent years, the release of three-dimensional simulation software for autonomous driving, such as Carla \cite{dosovitskiy2017carla} and CarSim \cite{benekohal1988carsim}, marks a shift from two-dimensional bird's eye view (BEV) simulations to more realistic three-dimensional scenarios. 
During simulation testing, experimenters need to create static scenes and incorporate dynamic elements such as traffic and pedestrians to build a comprehensive experimental scenario. 
Some of these scenarios are designed to challenge the performance and effectiveness of autonomous driving algorithms and are referred to as adverse scenarios.

Constructing static scenes in a three-dimensional model often requires experimenters to manually build three-dimensional model prefabs, place them in the scene, and configure their parameters and attributes to test the algorithm's performance under various conditions, which is both time-consuming and labor-intensive \cite{consumed}. 
Unfortunately, there is a scarcity of open-source adverse scenario sets for underground parking garages, which poses a challenge for testing autonomous vehicle parking tasks.

For Level 5 (L5) autonomous vehicles \cite{blog2021sae}, Automated Valet Parking (AVP) represents the final task in one driving journey. 
AVP allows passengers to leave the vehicle in a drop-off zone, such as at the entrance of a parking lot, and enables the vehicle to autonomously complete the subsequent parking maneuvers \cite{lan2023virtual}.
\begin{figure}[!ht] \centering
    \includegraphics[width=.45\textwidth]{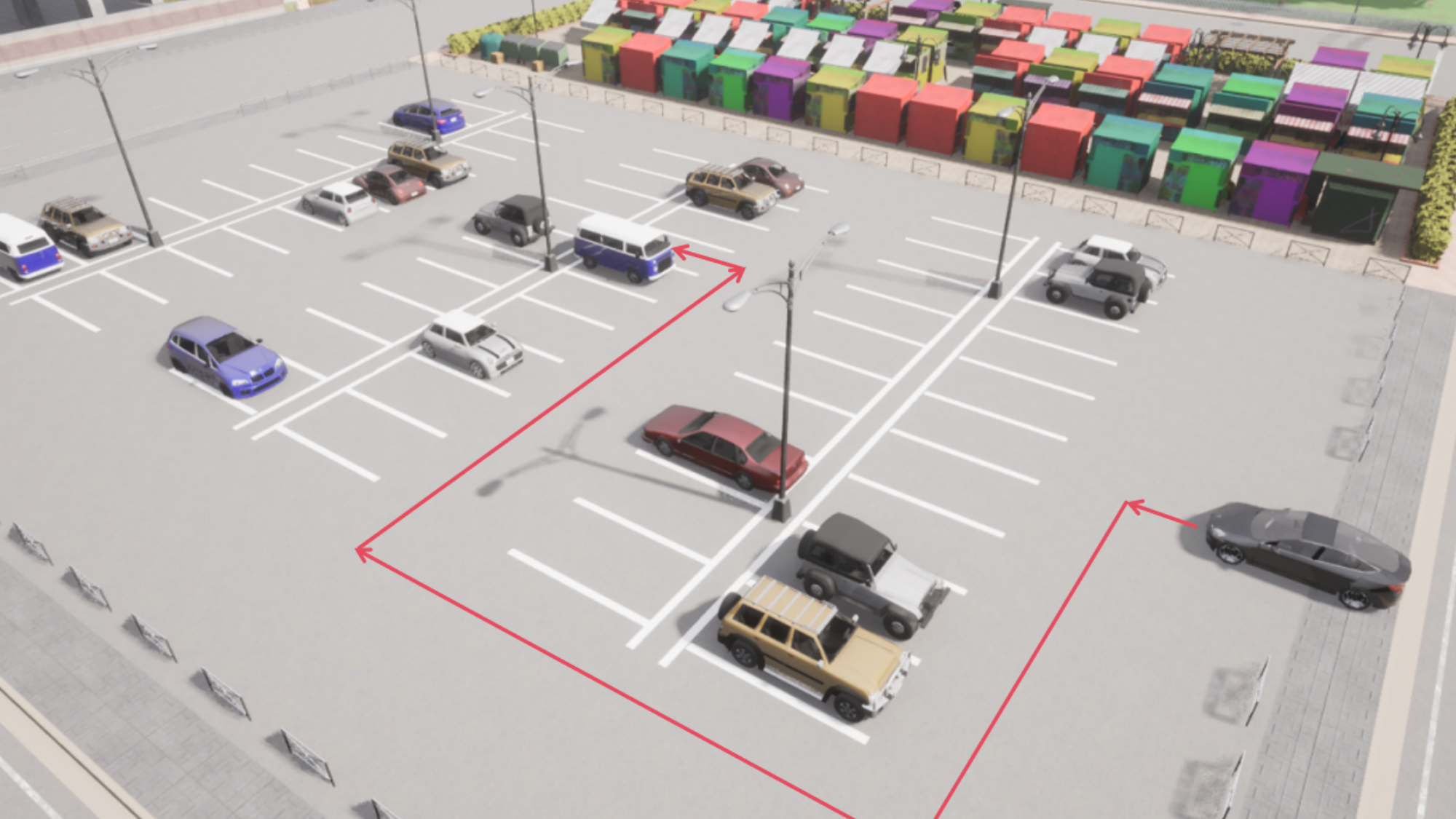}
    \caption{A example of automated valet parking. Red lines represent the predicted path of the car deployed with ADS.}\label{F:avp}
\end{figure}
For most ADS parking tasks, the real-world scenarios are typically underground parking garages (hereinafter referred to as "garages"). 
We aim to develop an automated construction framework to address the time and effort issues associated with manually constructing these scenarios. 
This framework will generate a corresponding set of three-dimensional adverse underground garage scenarios based on different input parameters and will be compatible with mainstream 3D simulation software. 
This approach seeks to resolve the current simulation testing issues of having few static scenes and manual scene construction's time-consuming, labor-intensive nature.

In this context, this paper proposes a reinforcement learning (RL) \cite{wiering2012reinforcement}-based procedural generation framework, using underground garage scenarios for AVP tasks as a representative example.
The framework will 
1) automate the construction and generation of simulated adverse underground garage scenarios in the Carla environment, 
2) utilize deep reinforcement learning to create diverse underground garage scenarios, and 
3) support customizable parameters and initial maps to alter the generated outcomes.

\subsection{Contributions}
The main contributions of this paper are as follows:

1. \textbf{New Automated Method for Generating Autonomous Driving Underground Garage Scenarios}: 
Under the proposed procedural generation framework, we can create diverse static underground garage scenarios. 
It significantly reduces the time experimenters spend on constructing static scenes. 
Additionally, these static scenes can be combined with dynamic elements to produce more challenging underground garage scenarios.

2. \textbf{High-Quality Encoding Using Reinforcement Learning}: 
The paper employs advanced reinforcement learning methods, designing environments, and reward functions for training to obtain higher-quality encodings. 
Since the reward functions are based on real-world rules \cite{lan2019simulated}, we can extract relatively optimal scenarios from the generated data. 
These scenarios make efficient use of space and closely resemble the distribution of roads and parking spaces in real underground garages, making them more applicable to real-world automated parking situations encountered by autonomous vehicles.

3. \textbf{Generation of Diverse Adverse Scenarios}: 
The encodings ultimately yield a variety of map models, forming a comprehensive map scenario library. 
We have designed rule-based validation metrics to differentiate the difficulty levels of various underground garages. 
These adverse scenarios of varying difficulty levels reduce the workload of experimenters in designing 3D scenes and provide more randomized test data, thereby enhancing the test coverage of algorithms.

\subsection{Outline}
This paper outlines the related work referenced or cited in the design of the procedural generation framework. 
Chapter 2 introduces the current state of research on static scene generation in simulation testing, the basic concepts of procedural generation, relevant methods in reinforcement learning, and the 3D autonomous driving simulation software Carla, including the file formats related to map generation. 
Chapter 3 elaborates on the system composition and core content of the framework: static underground garage scene generation based on deep reinforcement learning, divided into four parts: mathematical model construction, environment design, reward function design, and 3D map generation. 
Chapter 4 describes the experimental design for the deep reinforcement learning component of the framework, introduces two evaluation metrics—coverage and difficulty—and tests the planning algorithm, one of the core AVP algorithms, in Carla to verify the reliability of the difficulty metric for underground garages. 
Chapter 5 presents the experimental results designed in Chapter 4 and discusses these results. 
Chapter 6 summarizes the entire work and proposes future research directions.

\section{Related Work}
\label{sec:related_work}
\subsection{Static Scenes Generation for Simulation Testing}

The generation of simulation testing scenes for autonomous vehicles is an area still under exploration. 
After an extensive review of related literature, this paper focuses more on procedural model generation, which is commonly found in the construction of game elements within certain video games. 
Procedural Content Generation (PCG) refers to the creation of game content using algorithms with limited or indirect user input \cite{shaker2016procedural}. 
This paper primarily discusses a major research direction of PCG—generating different levels and scenes. 
PCG is frequently used in games, such as in the game "Minecraft," where PCG algorithms generate the entire block world and its contents. In the context of this paper, the game content is the 3D model of the underground garage, and the user inputs are the dimensions, shape, and other elements of the garage. 
Therefore, the problem of underground garage generation in this paper can be seen as a PCG problem within the transportation domain.

The research problem addressed in this paper is the automated creation of simulation testing scenes. 
Autonomous driving simulation testing scenes include static and dynamic scenes. 
Static scenes comprise a series of objects, such as road networks and static obstacles, that do not change positions during the simulation. 
On the other hand, dynamic scenes include a series of movable vehicles and pedestrians forming traffic flows. This paper primarily focuses on the generation of static scenes for testing.
The core content is the generation of the road network for the generation of static underground garage scenes.

\textbf{Rule-Based Road Network Generation}: 
Parish and Muller \cite{parish2001procedural} extended the L-system \cite{lindenmayer1968mathematical}—a system used to simulate the morphology of various organisms—to generate the road network in the CityEngine \footnote{https://www.esri.com/en-us/arcgis/products/esri-cityengine} urban generation system. 
Chen et al. \cite{chen2008interactive} proposed a procedural road network content generation method guided by underlying tensor fields, enabling intuitive user interaction.

\textbf{Search and Evolutionary Algorithm-Based Road Network Generation}: 
Ishhan Paranjape et al. developed the system CruzWay \cite{road-network2} to support the creation of test scenarios. 
CruzWay uses TownSim \cite{TownSim}, an agent-based urban evolution algorithm, to automate road network generation. 
The aforementioned works based on rules, search, and evolutionary algorithms \cite{lan2022time,lan2021learning,lan2021learning2} use urban semantic elements, such as elevation maps, water resource maps, and population density maps, as input to the program \cite{lan2022semantic}. 
These methods can ultimately generate large-scale urban transportation networks. However, the road network in this paper is smaller in scale compared to urban networks and has more restrictive rules, such as the absence of overly wide lanes and the presence of numerous orthogonal intersections. 
Alessio Gambi et al. used a combination of genetic algorithms and procedural content generation techniques to automatically create challenging simulation road networks for testing lane-keeping algorithms, resulting in more instances of lane departure compared to random testing \cite{road-network1}.
This work shares the same goal as this paper: providing test scenarios for autonomous driving system algorithms.

\textbf{Deep Learning-Based Road Network Generation}: 
Chu et al. proposed Neural Turtle Graphics (NTG), a method using neural networks to generate road networks represented by graphics. 
Owaki et al. introduced a new network model, RoadNetGAN \cite{RoadNetGAN}, an extension of NetGAN \cite{NetGAN}, which generates urban road networks by predicting road connections. 
These supervised learning methods use datasets for training and can generate road network structures similar to those in the datasets. The quality of these methods is determined by comparing the generated networks with those in the datasets. However, research focusing on underground garages as test scenarios is lacking, and there are insufficient datasets for evaluating the generated garbage results. 
This paper employs deep reinforcement learning for road network generation. This method does not rely on any dataset, which we believe provides higher coverage and randomness. Additionally, it can adapt to complex constraint rules and performs best in small to medium-sized scenarios.

\subsection{Reinforcement Learning}
Reinforcement Learning (RL) is a branch of machine learning that focuses on how agents take different actions in an environment to maximize rewards. 
RL comprises agents, environments, states, actions, and rewards. The agent is responsible for training and decision-making, while the environment includes state sets, action sets, and a reward function. 
States represent the data set in the environment, and the reward function specifies the numerical rewards the environment provides to the agent during the interaction. 
In an episode, the environment and the agent interact and progress together.
The interaction between the agent and the environment can be viewed as a finite Markov Decision Process (finite MDP), an efficient model for solving linear decision problems.

In this paper, finite MDPs for the reinforcement learning process are defined as follows: 
finite MDPs can be represented as a quadruple $M=\{S, A, P, R\}$, where $S$ is the finite state set, $A$ is the finite action set, $P : S\times R \times S \times A \to [0, 1]$ is the probability transition function, and $R : S \times A \to \mathcal{R}$ is the reward function. 
Here, we define the probability of transitioning from state $s$ to the next state $s'$ given action $a$ as $P(s', r|s, a)$, and the immediate reward obtained from this transition as $r_s^a$ \cite{ferns2003metrics}.

We define a policy $\pi$, with the mapping represented as $\pi: S\times A\to [0,1]$. 
We use $\pi(s)$ to denote the action $a$ in state $s$ under policy $\pi$. To evaluate the quality of a policy, we use the action-value function $q_\pi(s, a)$ to estimate the expected long-term cumulative reward of taking action $a$ in state $s$ under policy $\pi$. It is formally defined as:
\begin{equation}
q_\pi(s,a)=\mathbb{E}_\pi[\sum_{k=0}^\infty\gamma^kR_{t+k+1}| S_t=s, A_t=a]
\end{equation}
where $\gamma$ is the discount factor, $R_t$ is the reward at time step $t$, and $\Bbb{E}_\pi$ denotes the expectation with respect to policy $\pi$.

The goal of reinforcement learning is to find an optimal policy $\pi_*$ that maximizes the expected long-term discounted cumulative reward starting from any initial state $s\in S$:
\begin{equation}
    \pi_*=\operatorname*{argmax}_{\pi} \Bbb{E}_\pi [\sum_{t=0}^\infty \gamma^t R_t|s_0=s]
\end{equation}

\begin{figure} [!ht] \centering
    \includegraphics[width=0.7\linewidth,trim={20 30 20 20},clip]{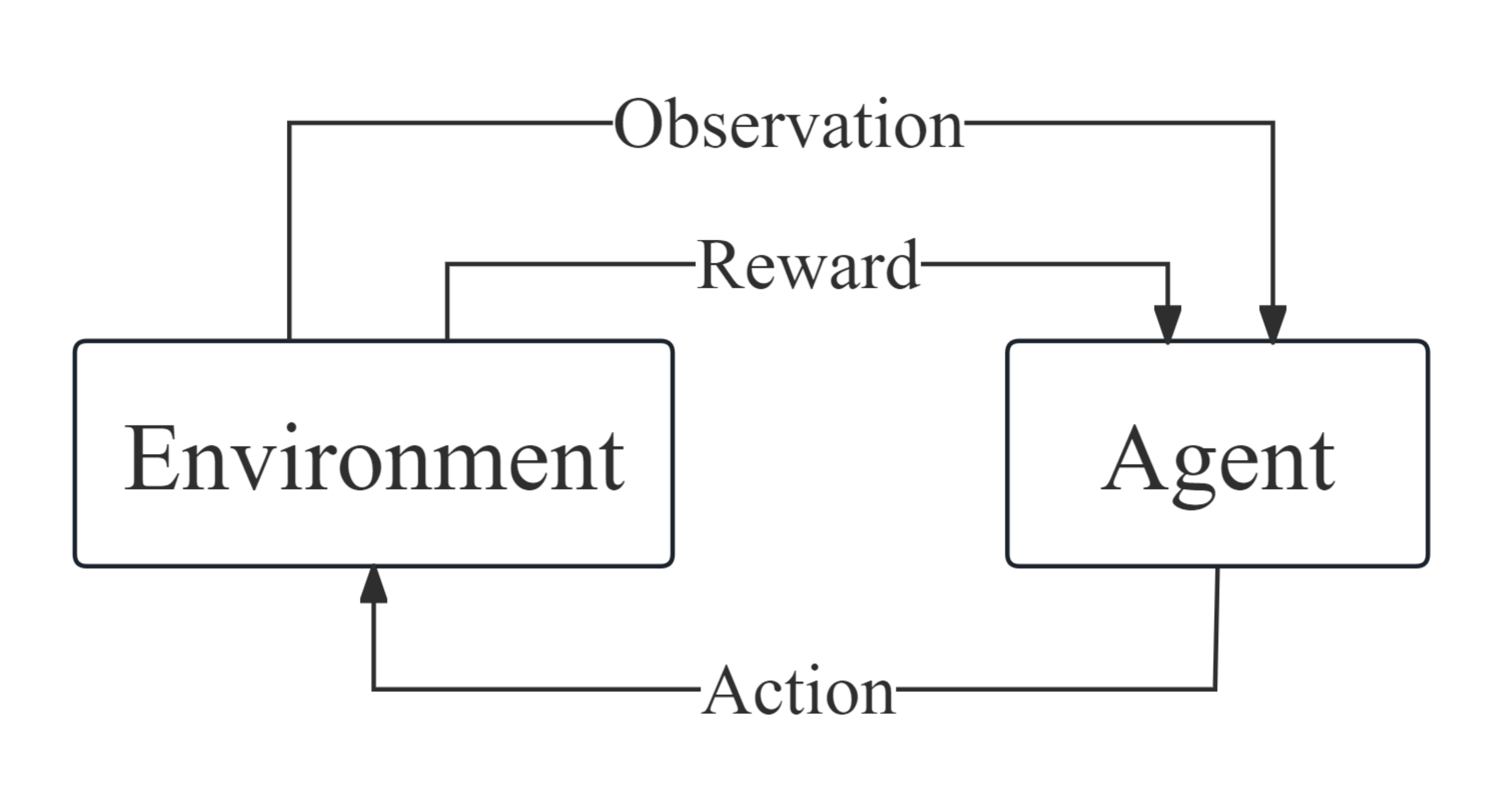}
    \caption{The framework of MDPs.}
    \label{F:MDPs}
\end{figure}

The interaction process between the agent and the environment is illustrated in \autoref{F:MDPs}. 
Initially, the environment provides the agent with the current state and reward. Based on the state information, the agent decides on an action, and upon receiving the action, the environment provides a new state and reward for the next cycle. 
This process continues until reaching a final state.

In each episode, the agent and the environment generate a finite Markov sequence:
{\small
\begin{equation}
    S_0,A_0, R_1, S_1, A_1, R_2, S_2, A_2, R_3,\dots, S_{T-1}, A_{T-1}, R_T
\end{equation}}
where $T$ denotes the termination time. The sum of all rewards $R_i$ yields the return $G_0$ for this episode:
\begin{equation}
    G_0 = R_1 + R_2 + \dots + R_{T} = \sum_{k=1}^T R_k
\end{equation}
We can update the action-value function $q(S, A)$ backward upon obtaining this sequence. 
The updated formula is as follows:
\begin{equation} \centering \small
\begin{split}
    q(S_t, A_t) = & \ q(S_t, A_t) + \alpha \times \big( R_{t+1} + \gamma \times q(S_{t+1}, A_{t+1}) \\
    & \ - q(S_t, A_t) \big)
\end{split}
\end{equation}
where $\alpha$ and $\gamma$ are adjustable parameters satisfying $0\le \alpha, \gamma \le 1$. Through continuous iteration, $q(s, a)$ will gradually converge towards the optimal value.

In recent years, with the rise of deep learning, the reliance on deep neural networks' powerful function approximation and representation learning capabilities has provided us with new tools to overcome high-dimensional problems \cite{lan2022vision}. 
The emergence of deep learning has had a significant impact on many fields of machine learning \cite{lan2022class,gao2021neat}, greatly enhancing the accuracy of tasks such as object detection \cite{lan2019evolving,lan2018real}, speech recognition, and language translation. 
Deep reinforcement learning algorithms are algorithms that combine reinforcement learning with deep learning. The Deep Q-Network (DQN) algorithm \cite{DQN} is a typical deep reinforcement learning algorithm. 
DQN addresses the instability of function approximation techniques in RL, demonstrating that RL agents can be trained solely based on reward signals from raw high-dimensional observations for the first time. 
DQN replaces the action-value function $q(s, a)$ in RL with a neural network $q(s, a, \theta_i)$ (also called the Q-network), where $\theta_i$ is the weight vector for the $i$-th iteration. 
By performing gradient descent on the Q-network, DQN continuously approximates the optimal policy.

\subsection{3D Autonomous Driving Simulation Software}
The Carla Simulator \cite{dosovitskiy2017carla} (hereinafter referred to as Carla) is an open-source simulation simulator developed by Intel Labs, Toyota Research Institute, and the Computer Vision Center, based on the Unreal Engine (UE). 
Carla is written in C++ and supports C++ and Python APIs. Compared to other commercial simulation software, Carla supports convenient custom map importing. 
Carla can be compiled and built directly from the source code, and supports map generation using files in Film Box and OpenDrive formats. 
Therefore, this paper selects Carla as the 3D autonomous driving simulation software for map integration.

\subsection{Map Formats}
FilmBox (FBX) \footnote{\url{https://www.autodesk.com/developer-network/platform-technologies/fbx-sdk-2020-0}} is a universal 3D model file format developed by Kaydara. 
Acquired by Autodesk in 2006, it has since become a primary exchange format for many 3D modeling tools. 
In Carla, FBX files are used to construct static scenes and provide volumes for collision detection. This paper uses the FBX Software Development Kit to generate FBX files.

ASAM OpenDrive (OpenDrive) \footnote{\url{https://www.asam.net/standards/detail/opendrive/}} uses Extensible Markup Language (XML) to describe road networks. 
OpenDrive can describe the geometry of roads, lanes, and related objects such as road signs and traffic lights. 
In the OpenDrive format, a complete road consists of many segments. 
Each road segment is represented by a \textless road\textgreater element, and the road reference line determines the centerline of the road in space, typically represented as a straight line, curve, or spiral. Intersections are represented by \textless junction\textgreater elements and can describe areas where three or more roads intersect. 
Intersections are defined as connecting three or more roads. 
In Carla, OpenDrive provides a list of waypoints, which can be used for path planning for autonomous vehicles. During point-to-point cruising tasks, vehicles can navigate by finding the closest path containing a series of waypoints. 
This paper uses the scenario generation Python library to generate OpenDrive files.

\section{Methodology}
\label{sec:methodology}

\begin{figure}[!ht]  \centering
    \begin{subfigure}[t]{.32\linewidth}  \centering
        \includegraphics[width=.95\textwidth,trim={10 0 10 0},clip]{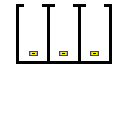}
        \caption{\textbf{Three-Stall Parking Space}: Three sequential car stalls are in one square block.}\label{F:three-stalls-a}
    \end{subfigure}
    \begin{subfigure}[t]{.32\linewidth}  \centering
        \includegraphics[width=.95\textwidth,trim={10 0 10 0},clip]{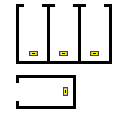}
        \caption{\textbf{Four-Stall Parking Space}: Three sequential car stalls and one alone stall are in one square block. }\label{F:three-stalls-b}
    \end{subfigure}
    \begin{subfigure}[t]{.32\linewidth}  \centering
        \includegraphics[width=.95\textwidth,trim={10 0 10 0},clip]{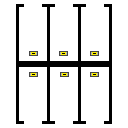}
        \caption{\textbf{Six-Stall Parking Space}: Six car stalls are in one square block.}\label{F:three-stalls-c}
    \end{subfigure}
    \caption{Three combinations of vertical parking space.}\label{F:three-stalls}
\end{figure}

To better address the issue of generating underground parking garage scenes, we outline the following requirements for the generated static scenes:

\begin{enumerate} [leftmargin=*]
    \item The parking garage should be a simple single-level structure without interior walls.
    \item Only double-lane road segments in horizontal and vertical directions should exist within the parking garage.
    \item There should be only one entrance and one exit for the parking garage, and they should be connected by the road network.
    \item All parking spaces should be vertical parking spaces.
    \item Following the "Garage Architectural Design Code", vertical parking spaces should appear in groups of three, four, and six in the parking garage, as illustrated in \autoref{F:three-stalls}.
\end{enumerate}

The generation of adverse scenarios in underground parking lots is divided into several steps as follows:
\begin{itemize}
    \item Firstly, a two-dimensional grid map is generated to represent the planar structure of the parking lot. 
    Each grid represents a minimum separable block within the parking lot, such as roadblocks and parking space blocks. The grid map is encoded using a matrix.
    \item Subsequently, each block of the encoding matrix is mapped to a three-dimensional scene and additional static obstacles, such as pillars and barriers, are incorporated. This step generates files representing the three-dimensional models.
    \item Finally, the roadblocks of the encoding matrix are transformed into road network format information. This step produces files representing the road network.
\end{itemize}
The proposed procedural generation method consists of four parts: mathematical model construction, environment design, reward function design, and three-dimensional map model generation. 
The first part addresses the parking lot generation problem requirements and constructs a mathematical model accordingly. 
The second part translates the mathematical model into a problem within the reinforcement learning framework and designs its environment. 
The third part designs appropriate reward functions to ensure that the training results meet the requirements of the first part and make the parking lot as usable as possible. 
The final part uses the encoding matrix as input to generate FBX and XODR files, which are then imported into Carla.
\begin{figure*}[!ht] \centering
    \includegraphics[width=0.97\textwidth,trim={0 10 0 10},clip]{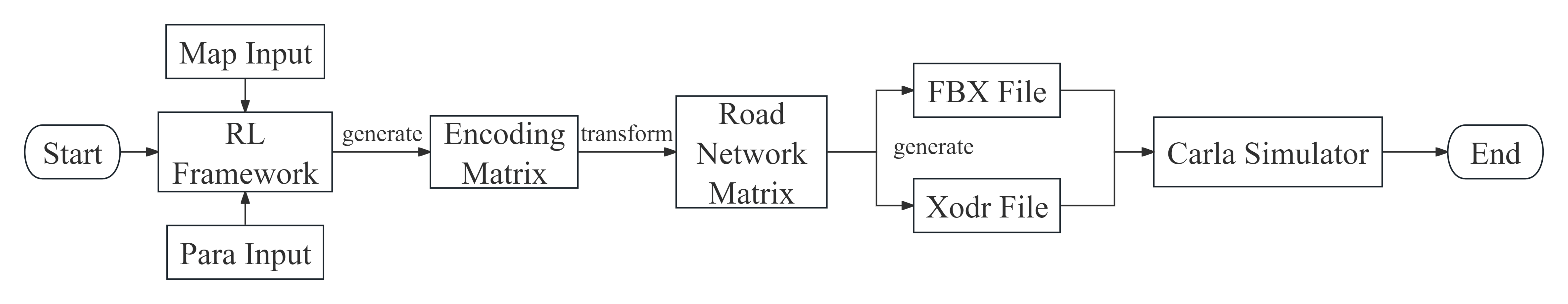}
    \caption{The PCG Framework. It starts at the RL framework. RL framework needs the input of one initial map and parameters. The train process of RL framework generates the encoding matrix set and they can be transformed into the road network matrix set. The road network matrix set will generate fbx files and xodr files, which can import into Carla simulator as the 3D map. }\label{F:framework}
\end{figure*}
The entire execution process of the method is illustrated in \autoref{F:framework}. 
The RL framework supports user input of custom parameters and maps. After training, it generates a set of encoding matrices. These encoding matrices can be converted into road network matrices, which are then used to generate corresponding FBX and Xodr files. 
These files can be imported into the Carla simulator to support further training.

\subsection{Mathematical Model Construction}
We classify the minimum separable blocks in the construction of the underground garage. The categories are presented in \autoref{T:block-code}.
\begin{table}[!ht] \centering 
    \caption{The taxonomy of blocks in the underground garage.}
    \label{T:block-code}
    \begin{tabular}{lc|r} \toprule
        Block Name & Code & \multicolumn{1}{r}{Variant} \\ \midrule
        Free Block & 0 & $\mathrm{FREE\ {BLOCK}}$ \\
        Road Block & 1 & $\mathrm{ROAD\ {BLOCK}}$ \\
        Obstacle Block & 2 & $\mathrm{OBSTACLE\ {BLOCK}}$ \\
        Obst-Three-Stall Block & 3 & $\mathrm{OBSTACLE\ STALL3\ {BLOCK}}$ \\
        Three-Stall Block & 4 & $\mathrm{STALL3\ {BLOCK}}$ \\
        Four-Stall Block & 5 & $\mathrm{STALL4\ {BLOCK}}$ \\
        Six-Stall Block & 6 & $\mathrm{STALL6\ {BLOCK}}$ \\
        Entrance Block & 7 & $\mathrm{ENTRANCE\ {BLOCK}}$ \\
        Exit Block & 8 & $\mathrm{EXIT\ {BLOCK}}$ \\
        Obst-Four-Stall Block & 9 & $\mathrm{OBSTACLE\ STALL4\ {BLOCK}}$ \\  \bottomrule
    \end{tabular}
\end{table}
Each number represents a block type. Thus, we can utilize any of these encoding matrices to represent the planar structure of an underground garage. 
Among these, adjacent roadblocks are interconnected, and we aim for accessibility from any roadblock to any other. Entrances and exits are linked to roadblocks, ensuring travel from an entrance to an exit. 
There are five types of parking spaces: three-stall, four-stall, six-stall, obstructed three-stall (obst-three-stall), and obstructed four-stall (obst-four-stall). 
The obstructed three-stall and obstructed four-stall blocks are necessarily connected to obstructed blocks. 
This handling ensures that parking spaces are located along the sides of roads to accommodate vehicles.

\begin{figure}[!ht] \centering
    \begin{subfigure}[t]{.48\linewidth}   \centering
        \includegraphics[width=1\textwidth]{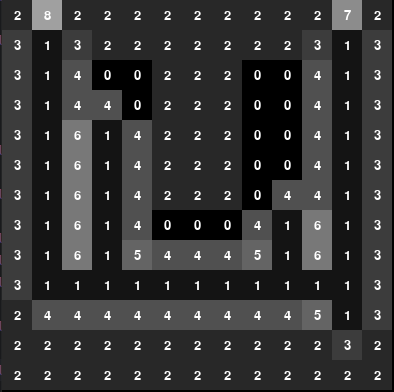}
        \caption{Code Matrix}\label{F:code}
    \end{subfigure}
    \begin{subfigure}[t]{.48\linewidth}   \centering
        \includegraphics[width=1\textwidth]{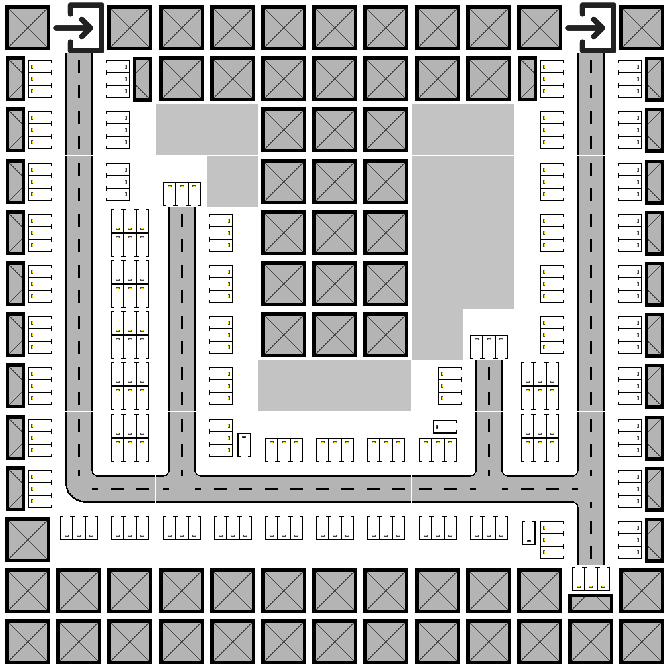}
        \caption{Grid Map}\label{F:structure}
    \end{subfigure}
    \caption{The transformation from encoding matrix to grid map.}\label{F:code-struct}
\end{figure}

\begin{figure*}[!ht] \centering
    \includegraphics[width=.95\textwidth,trim={10 10 10 10},clip]{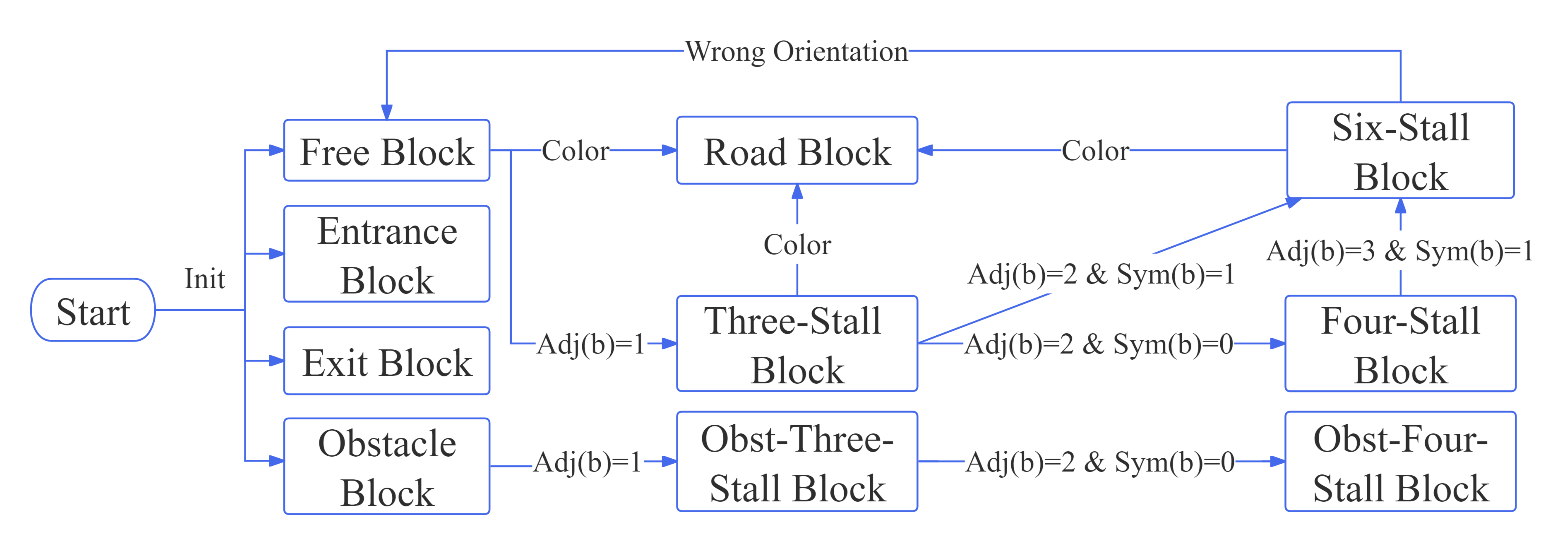}
    \caption{The state machine for block $b$. The initial map decides every concrete initial block type. The initial state of one block is one of four block types: free block, entrance block, exit block, and obstacle block. The equation($b$) shows the condition of transformation. The transformation occurs if the equation($b$) is true. '\&' symbols 'and'. The wrong orientation means the environment needs the north-south six-stall block but the direction of the six-stall block is east-west. The direction of the six-stall block depends on the positions of the adjacent road blocks.}\label{F:block-state-machine}
\end{figure*}

This study employs the deep reinforcement learning approach to generate the encoding matrix, necessitating the definition of various metrics within the reinforcement learning framework for this problem. 
To this end, we transform the encoding matrix generation problem into a dynamic problem solvable by reinforcement learning. 
We assume the presence of a car, which starts from the entrance block of the garage and can move to adjacent blocks in each step. During its progression, the car "colors" the passed blocks as roadblocks and colors the adjacent blocks of the colored roadblocks. The coloring rules for individual blocks $b$ are governed by a state machine, as illustrated in \autoref{F:block-state-machine}. 
Blocks are initially transitioned based on the input map into free blocks, obstacle blocks, entrance blocks, or exit blocks. 
While entrance and exit blocks remain unchanged, the other two types of blocks undergo different transitions. These transitions are primarily classified into two categories: coloring and changes in the states of adjacent blocks. When the car is at a particular block, a \textbf{coloring} action occurs. 
We define the function $Adj(b)$ to represent the number of roadblocks among the four adjacent blocks of block $b$, with the formula defined as:
\begin{equation}
    \begin{aligned}
        Adj(b)=& \ |\{b'|\mathrm{type}_{b'}=\mathrm{ROAD\ BLOCK},\\
            & \ b'\in\{b_{left}, b_{top}, b_{right}, b_{down}\}\}|
    \end{aligned} \label{eq:xxx}
\end{equation}
where $b_{left}$, $b_{top}$, $b_{right}$, and $b_{down}$ represent the blocks to the left, top, right, and bottom of block $b$, respectively. 
Additionally, we define $Sym(b)$ as an indicator of whether the adjacent block set of block $b$ contains symmetric roadblocks. Specifically, when both $b_{left}$ and $b_{right}$ are roadblocks, or when both $b_{top}$ and $b_{down}$ are roadblocks, $Sym(b)$ is expressed as $1$; otherwise, it is $0$.

It is noteworthy that this study introduces specific rules for the blocks with six parking spaces.
All six parking spaces in a single block must face the same direction, either north-south or east-west. 
This direction is determined by the environment at the beginning of each episode.
Eventually, we obtain an encoding matrix colored by the car's movements. This problem can be addressed using reinforcement learning techniques. 
The transformed problem is defined as follows: the car is situated in a parking garage environment, initially containing only one entrance and one exit (non-overlapping).
The agent can control the car to move to adjacent blocks and color them, aiming to generate a parking garage that meets the constraints as much as possible while remaining usable. 
We need to design the state space, action space, and reward function for the environment, which will be detailed in the next subsection. 
After each episode, reinforcement learning will produce a complete encoding matrix for the parking garage. The collection of encoding matrices generated across all episodes constitutes the output of this reinforcement learning iteration.

\begin{figure*}[!ht]  \centering
    \includegraphics[width=.75\textwidth,trim={10 10 10 20},clip]{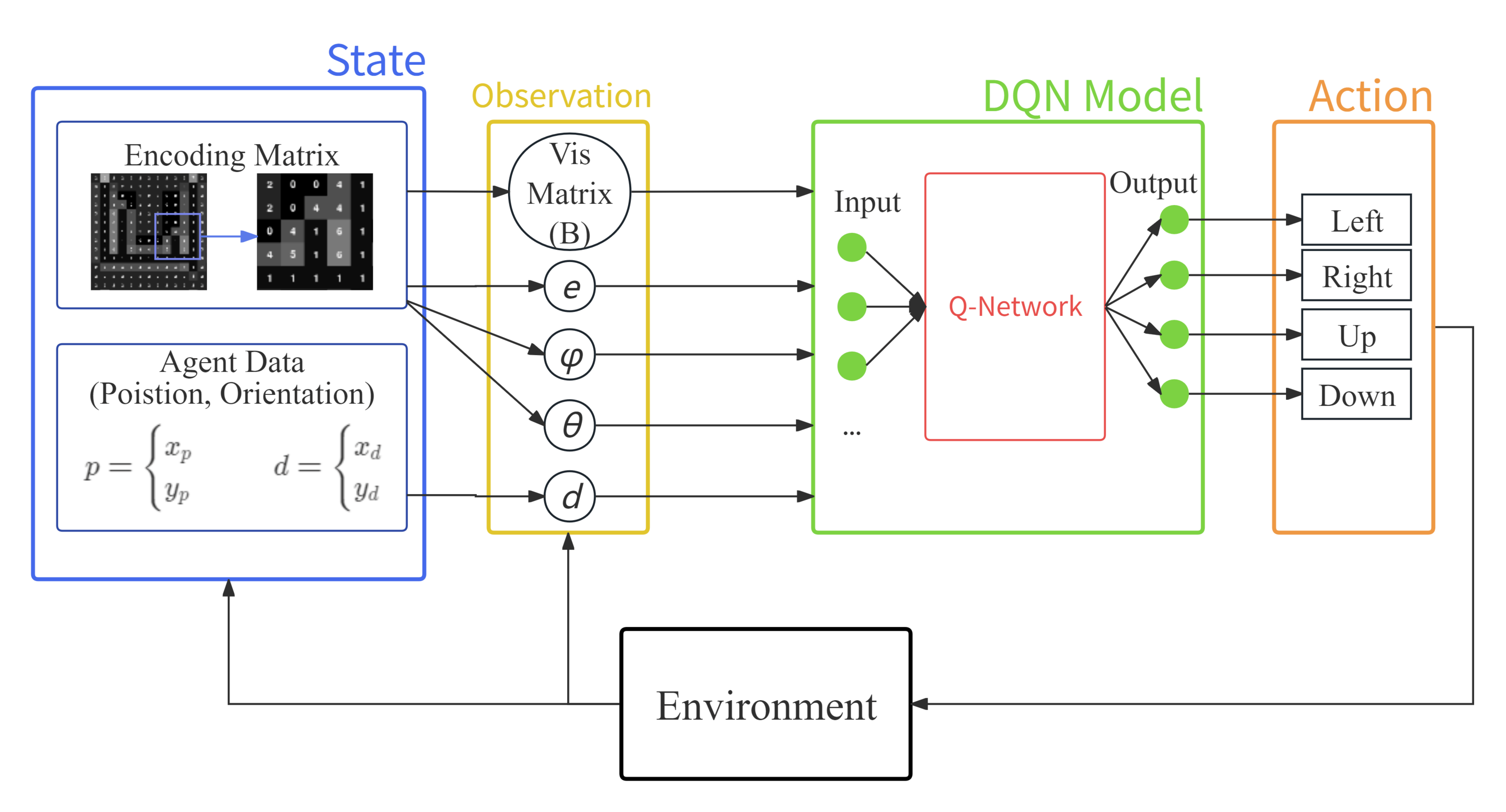}
    \caption{The DQN reinforcement learning framework, which contains four parts: 
    blue part for state containing encoding matrix and agent data; 
    yellow part for observation containing visibility matrix $\mathrm{B}$, error index $\mathrm{e}$, coverage rate $ \varphi $, connectivity index $ \theta $, and car orientation $ \mathrm{d} $; 
    green part for reinforcement learning model;
    brown part for four actions.} \label{F:rl-architecture.png}
\end{figure*}

\subsection{Environment Design}
The core of environment design lies in defining the state space $ S $ and action space $ A $. 
The environment in this paper primarily consists of an encoding matrix and data representing the car. The encoding matrix indicates the range of movements available to the car in the environment, while the car data stores the position and orientation of the car within the encoding matrix.
The actions of the car are defined as moving left, right, up, or down, and the action set $ A $ is defined as follows:
\begin{equation} \centering
    A=\{\text{Left}, \text{Right}, \text{Up}, \text{Down}\}
\end{equation}
At each step, the car can move to an adjacent block.
The environment in this paper is designed as a "partially observable environment." 
In this environment, \textbf{observations} are used instead of the \textbf{states} mentioned in the MDP framework. This means that the agent receives \textbf{observations} from the environment rather than \textbf{states}.
In a partially observable environment, observations primarily consist of a visibility matrix $ \mathrm{B} $, error index $ \mathrm{e} $, coverage rate $ \varphi $, connectivity index $ \theta $, and car orientation $ \mathrm{d} $. 
In this environment, we assume that the agent cannot access the entire structure of the encoding matrix or the position of the controlled car to make decisions. Instead, the agent processes features extracted from the encoding matrix to make decisions.
These features include the visibility matrix, error index, coverage rate, and connectivity index. 
The visibility matrix refers to a sub-matrix of the encoding matrix with the same width and height, centered around the car. It simulates the observable range of a human driver in a parking garage. 
The error index indicates the number of times the agent violates constraints. 
If the error index exceeds the maximum allowable value, the current episode ends directly. 
The coverage rate represents the percentage of colored blocks in the garage relative to the initial number of uncolored blocks. It evaluates the extent to which the agent has colored the garage. 
The connectivity index is a Boolean value indicating whether the entrance and exit are connected. 
The garage is considered usable only when the car has reached the exit, indicating that the starting point and the exit are connected.

The observation space in the constrained environment is smaller than the state space, which speeds up training but may lead to the loss of some structural information about the garage.

\subsection{Reward Function Design}
The design of the reward function is a crucial aspect of reinforcement learning, as it often determines the quality of training outcomes. 
In this study, the reward function is designed with reference to national parking garage design standards to closely approximate real-world garage design rules. Positive rewards indicate gains, while negative rewards imply losses. Larger rewards encourage actions, while smaller rewards discourage actions. 
The reward function is based on the rules of the environment, which are categorized into constraint rules and utility rules.

Constraint rules primarily restrict the agent's movement while ensuring connectivity between the entrance and exit roads of the garage. 
Utility rules focus on the user experience and efficiency of garage usage, aiming to maximize the number of parking spaces, increase the percentage of parking area in the garage, reduce the number of intersections, and improve traffic efficiency. Additionally, utility rules contribute to evaluating the quality of a garage, playing a crucial role in subsequent garage validation.

The reward function $ R(t) $ is defined as follows:
\begin{equation}
    R(t) = k_c R_c(t) + k_u R_u(t)
\end{equation}
where $ R_c $ represents the reward function for constraint rules, $ R_u $ represents the reward function for utility rules, and $ k_c $ and $ k_u $ are parameters ranging from 0 to 1.
The constraint rules mainly consist of road network constraints, obstacle collision constraints, entrance-exit connectivity constraints, and backward movement constraints.

\begin{figure}[!ht]  \centering
    \begin{subfigure}[t]{.47\linewidth}
        \centering
        \includegraphics[width=1\textwidth]{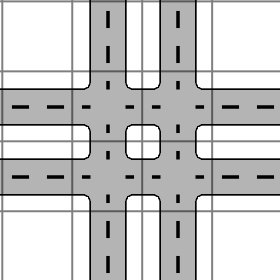}
        \caption{\textbf{$2\times 2$ Road Block Set}: Four adjacent road blocks compose a square which violates the road network rule in the real world.}\label{F:constraint1}
    \end{subfigure}
    \begin{subfigure}[t]{.47\linewidth}
        \centering
        \includegraphics[width=1\textwidth]{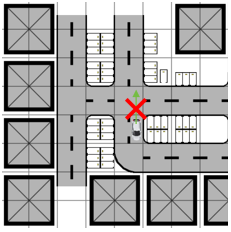}
        \caption{\textbf{Backward Movement Penalty}: the agent will obtain penalty if it turns back.}\label{F:constraint2}
    \end{subfigure}
    \caption{Two examples of constraint rule.}\label{F:constraint}
\end{figure}

\textbf{Road Network Constraint Rule}. 
Vehicles are prohibited from coloring blocks that form a 2×2 grid of roadblocks. Such a configuration is unrealistic, as depicted in \autoref{F:constraint1}.

\textbf{Obstacle Collision Constraint Rule}.
Vehicles receive penalties for colliding with obstacles.

\textbf{Entrance-Exit Connectivity Constraint Rule}.
The roads colored by the vehicle should connect the entrance and exit points; otherwise, the garage cannot function properly for parking.

\textbf{Backward Movement Constraint Rule}.
Vehicles are restricted from moving backward arbitrarily. Instead, they should prioritize forward, left turn, and corresponding left turn actions, as illustrated in \autoref{F:constraint2}. Penalties are imposed when vehicles move backward.

It is worth noting that the calculation of the error-index mentioned earlier does not consider violations of the backward movement constraint. 
This is because, in specific situations, vehicles may need to move backward to achieve higher rewards.
For instance, moving backward might be a less penalized action compared to crashing into an obstacle.

The utility rules mainly include the parking space quantity utility rule, road length utility rule, and intersection utility rule. 
Each of these rules provides a performance value based on the current state of the garage. 
When the agent takes action and changes the state of the garage, the performance value also changes accordingly, with the difference representing the reward for the current action.

\textbf{Parking Space Quantity Utility Rule}.
After each action, this rule evaluates the number of parking spaces in the garage. It rewards an increase in the number of parking spaces and penalizes a decrease.

\textbf{Road Length Utility Rule}.
This rule calculates the distribution of road lengths in the garage and computes a weighted performance value based on it. It aims to optimize the layout of the roads within the garage.

\textbf{Intersection Utility Rule}.
The rule calculates the distribution of different intersections within the garage and computes a weighted performance value based on it. 

The calculation of these utilities allows the reinforcement learning agent to evaluate the effectiveness of its actions and make decisions accordingly during the training process.
The reward for the utility rule part can be expressed as:
\begin{equation}
    \begin{split}
        R_u(t) = & \ w_s(P_s(t) - P_s(t-1)) \\
                 & \ + w_r(P_r(t) - P_r(t-1)) \\
                 & \ + w_c(P_c(t) - P_c(t-1)) \\
    \end{split}
\end{equation}
Here, $ P_s $, $ P_r $, and $ P_c $ represent the performance values of the parking space quantity utility rule, road length utility rule, and intersection utility rule, respectively. $ w_s $, $ w_r $, and $ w_c $ are adjustable parameters.

We define $ X_t $ as the random variable representing the road length at time $ t $, and $ Y_t $ as the random variable representing the number of intersections at time $ t $. Then,
\begin{equation}
P_r(t) = \sum_{i=0}^{c_{X_t}} P(X=x_i) w^{X_t}_{i}
\end{equation}
\begin{equation}
P_c(t) = \sum_{i=0}^{c_{Y_t}} P(Y=y_i) w^{Y_t}_{i}
\end{equation}
Here, $ w^{X_t}_{i} $ and $ w^{Y_t}_{i} $ are adjustable parameters, and $ c_{X_t} $ and $ c_{Y_t} $ represent the possible values of $ X_t $ and $ Y_t $ respectively. $ P(X=x_i) $ and $ P(Y=y_i) $ denote the probabilities associated with each value $ x_i $ and $ y_i $ of $ X_t $ and $ Y_t $ respectively.

\subsection{3D Maps Generation}

This section will explain how to convert the encoding matrix into a road network matrix and generate FBX and OpenDrive files. 
These files constitute the 3D map, which will be imported into Carla to ultimately generate the 3D garage.

We stipulate that each element in the encoding matrix has a length and width of 9 meters.

We convert the encoding matrix into a road network matrix. Each element in the road network matrix contains the current block type, coordinates, and orientation angle. 
The block type corresponds to our pre-built 3D models, the coordinates indicate the position of the pre-built 3D model in 3D space, and the orientation angle indicates how much the pre-built 3D model should be rotated. 
Here, we divide the roadblocks into four types: straight road, turning road, T-intersection, and crossroad, as shown in the figure below.

\begin{figure}[htb]
    \centering
    \begin{subfigure}[t]{.23\linewidth}
        \centering
        \includegraphics[width=0.8\textwidth]{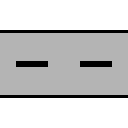}
        \caption{Straight Road}\label{F:lane-straight}
    \end{subfigure}
    \begin{subfigure}[t]{.23\linewidth}
        \centering
        \includegraphics[width=0.8\textwidth]{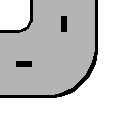}
        \caption{Curve Road}\label{F:lane2}
    \end{subfigure}
    \begin{subfigure}[t]{.23\linewidth}
        \centering
        \includegraphics[width=0.8\textwidth]{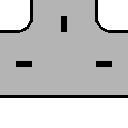}
        \caption{T-Junction Road}\label{F:lane3}
    \end{subfigure}
    \begin{subfigure}[t]{.23\linewidth}
        \centering
        \includegraphics[width=0.8\textwidth]{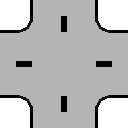}
        \caption{Cross Road}\label{F:lane4}
    \end{subfigure}
    \caption{The common four road types.}\label{F:lane-type}
\end{figure}

Once we obtain the road network matrix, we can correspondingly use pre-built block model files to combine and generate the entire three-dimensional floor model of the garage.

We extract all road information from the road network matrix to generate the OpenDrive file. 
Here, we categorize straight roads and turning roads into one category, defined as \textless road\textgreater in the OpenDrive format, while T-junction road and crossroad are grouped into another category, defined as \textless junction\textgreater in the OpenDrive format. 
For two adjacent \textless road\textgreater segments, we directly connect them. For \textless junction\textgreater segments, we connect them to the adjacent \textless road\textgreater segments. 
There might be cases where two T-intersections are adjacent. 
In such cases, we add an implicit road segment between the two intersections to connect them.
\section{Experiments}
\label{sec:experiments}
\subsection{Environmental Setup}
We set up the environment for our experiment as blow \autoref{T:environment}
\begin{table}[!ht] \centering
    \caption{Description for the environment parameter.}
    \label{T:environment}
    \setlength{\tabcolsep}{0.3cm}
    \begin{tabular}{lc}   \toprule
        Parameter & Description \\  \midrule
        Operating System          & Windows11 \\
        CPU          & Intel(R) Core(TM) i7-9700K CPU @ 3.60GHz\\
        GPU          & NVIDIA GeForce GTX 1080\\
        Memory             & 16G\\
        Python Version          & python3.7/python3.8 \\
        Carla Version        & 0.9.15 source build version \\
        Unreal Version       & 4.26.7 source build version \\ \bottomrule
    \end{tabular}
\end{table}
The reinforcement learning environment in this paper is custom-built using stable-baselines \cite{stable-baselines3}, and the algorithm employs the DQN algorithm encoded in stable-baselines. 
FBX files are generated using the FBX SDK provided by Autodesk,
while OpenDrive files are generated using the Python third-party library scenario generation. 
All initial data for the parking lots are manually authored.

\subsection{Assessment Metrics}
Carla provides the shortest feasible topological path $\mathcal{P}$ from any starting point to the destination. 
Path $\mathcal{P}$ consists of a series of interconnected waypoints. We define the drivable path in the parking lot as the shortest path from the entrance to any parking space along the roads. 
We establish two metrics to evaluate the diversity of generated parking lots: coverage $\delta$ and difficulty $\lambda$.
Coverage assesses the richness of drivable paths produced by the parking lot generation. Difficulty evaluates the complexity of drivable paths generated by the parking lot. Both coverage and difficulty are directly computed from the encoding matrix. 
For computational convenience, we approximate the coverage of a parking lot as a function of the remaining vacant blocks within the parking lot. The difficulty of a parking lot is approximated as a function of the expected road length and intersection count. 
This approximation corresponds to the utility rules defined in Section 3 regarding the reward function design.

We define the coverage $\delta$ of a parking lot $G$ as:
\begin{equation}
    \delta = 1 - \frac{|\{p | \text{type}_p = \text{FREE BLOCK}, p \in S_G\}|}{|\{p | \text{type}_p = \text{FREE BLOCK}, p \in S_{G'}\}|}
\end{equation}
where $S_G$ represents the set of blocks in parking lot $G$, and $G'$ denotes the initial structure of parking lot $G$. 

Let $X$ be the random variable representing road length, with minimum value $l_{min}=2$ and maximum value $l_{max}=6$. 
Similarly, let $Y$ be the random variable representing the number of road intersections, with minimum value $c_{min}=2$ and maximum value $c_{max}=4$. Normalizing the expectations, we obtain:
\begin{equation}
    n_1 = \frac{\mathbb{E}(X) - l_{min}}{l_{max} - l_{min}}
\end{equation}
\begin{equation}
    n_2 = \frac{\mathbb{E}(Y) - c_{min}}{c_{max} - c_{min}}
\end{equation}
Based on experience and utility rules, we consider longer road lengths and a higher number of intersections as simpler. Therefore, we define the difficulty $\lambda$ of a parking lot $G$ as:
\begin{equation}
    \lambda = w_1(1 - n_1) + w_2(1 - n_2)
\end{equation}
where $w_1$ and $w_2$ are parameters, with $w_1 = 0.33$ and $w_2 = 0.67$.

\subsection{RL Experiments}
\begin{figure*}[htb]
    \centering
    \begin{subfigure}[t]{.19\linewidth}
        \centering
        \includegraphics[width=1\textwidth]{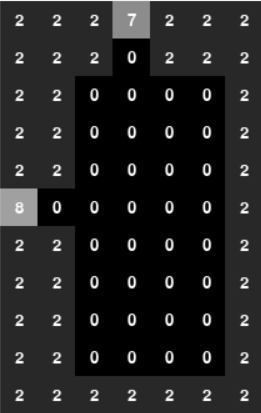}
        \caption{$11\times 7$ Garage}\label{F:origin-m3}
    \end{subfigure}
    \begin{subfigure}[t]{.3\linewidth}
        \centering
        \includegraphics[width=1\textwidth]{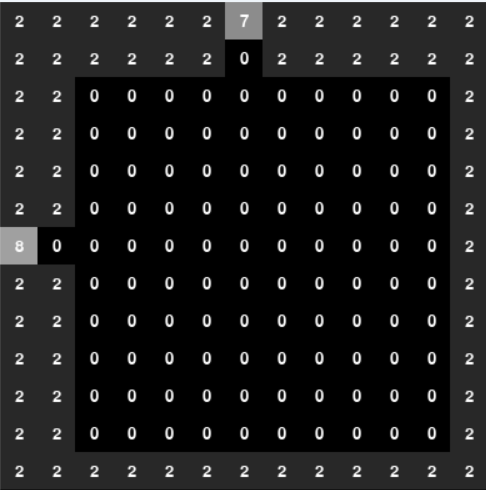}
        \caption{$13\times 13$ Garage without Corresponding Entrance and Exit}\label{F:origin-m4}
    \end{subfigure}
    \begin{subfigure}[t]{.3\linewidth}
        \centering
        \includegraphics[width=1\textwidth]{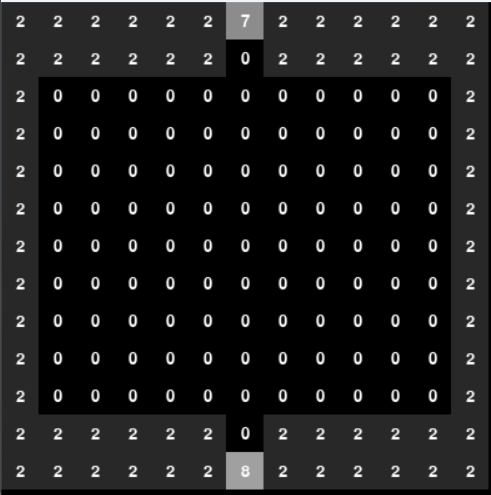}
        \caption{$13\times 13$ Garage with Corresponding Entrance and Exit}\label{F:origin-m5}
    \end{subfigure}
    \begin{subfigure}[t]{.3\linewidth}
        \centering
        \includegraphics[width=1\textwidth]{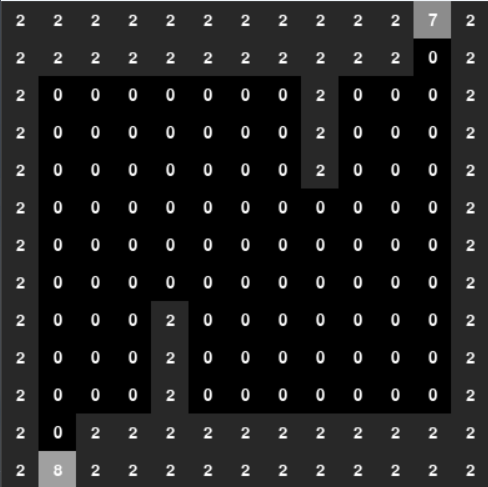}
        \caption{$13\times 13$ S-shaped Garage}\label{F:origin-m6}
    \end{subfigure}
    \begin{subfigure}[t]{.3\linewidth}
        \centering
        \includegraphics[width=1\textwidth]{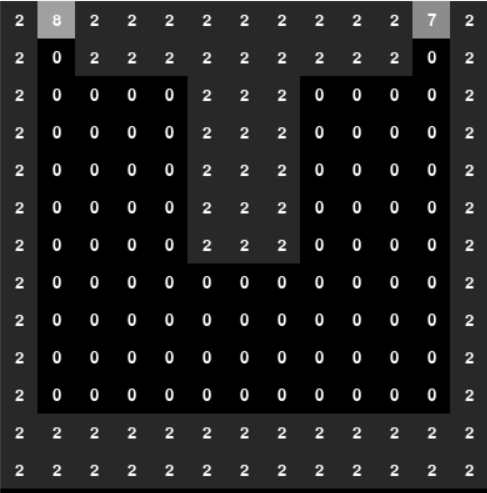}
        \caption{$13\times 13$ U-shaped Garage}\label{F:origin-m7}
    \end{subfigure}
    \caption{Five encoding matrices for initial maps.}\label{F:origin-map}
\end{figure*}

RL Experiments focus on generating encoding scripts using the DQN algorithm with five maps of different sizes.
As shown in \autoref{F:origin-map}, the first three maps are relatively regular, with rectangular vacant blocks in the center and obstacles at the edges, except for the entrance and exit. 
Map from \autoref{F:origin-m3} is smaller, while the others are larger. 
Maps from \autoref{F:origin-m4} and \autoref{F:origin-m5} have different exit positions. 
Since we encourage the creation of parking lots with multiple straight roads in reinforcement learning, the map from \autoref{F:origin-m5} is expected to be easier to generate usable parking lots.
The relatively challenging maps are from \autoref{F:origin-m6} and \autoref{F:origin-m7}, where obstacles obstruct the path to the destination, making it more difficult to find a viable route.
We trained each map five times, with a total of 2,000,000 timestamps per training run (timestamps represent the interactions between the agent and the environment), and plotted the convergence curves of their rewards. 
We also extracted all available garage structures from the training results for each map and plotted heat maps of coverage and difficulty distributions.

\subsection{Garage Difficulty Verification Experiments}

\begin{figure}[!ht]  \centering
    \begin{subfigure}[t]{.48\linewidth}
        \centering
        \includegraphics[width=1\textwidth]{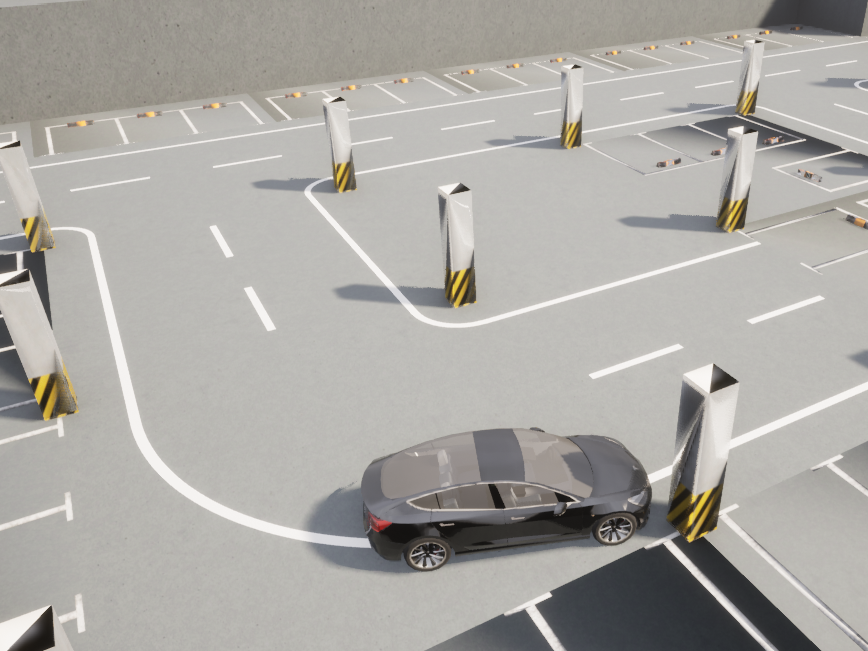}
        \caption{\textbf{Collision}: the vehicle is collided with a pillar.}\label{F:fail1}
    \end{subfigure}
    \begin{subfigure}[t]{.48\linewidth}
        \centering
        \includegraphics[width=1\textwidth]{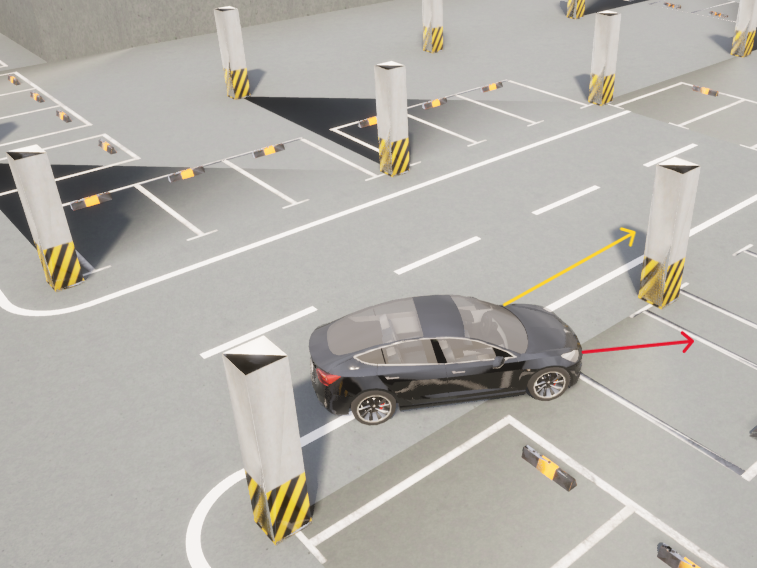}
        \caption{\textbf{Deadlock}: the vehicle is deviated from yellow trajectory. Red arrow represents the wrong direction.}\label{F:fail2}
    \end{subfigure}
    \caption{Failure cases for garage difficulty verification experiments.}\label{F:constraint}
\end{figure}

We designed an experiment to test one of the core algorithms of the AVP, the cruise algorithm. 
For this purpose, we used the garages generated in the previous experiment in Carla for automated driving cruise testing. We set the starting point and the finishing point in the garage, and the autonomous driving vehicle needs to depart from the starting point and finally reach the endpoint.
The vehicle may encounter the following situations during the journey: collision with parked vehicles, collision with walls, and algorithmic exceptions leading to deadlock. These situations will result in task failure. 
To simulate the parking task of the AVP in a real parking lot, we set the starting point as the entrance of the garage and the endpoint as the road edge closest to any parking space. 
This setting is consistent with the feasible driving path mentioned earlier. 
The vehicle has three failure cases: collision, timeout, and deadlock. Collision mainly occurs when the vehicle collides with the boundary walls and pillars, as shown in \autoref{F:fail1}; 
Timeout occurs when the vehicle fails to reach the endpoint within the predetermined time. 
Deadlock refers to algorithmic problems, such as deviation from the predetermined trajectory, and the vehicle cannot reach the specified position, as shown in \autoref{F:fail2}.

We selected the cruise algorithm based on the deep reinforcement learning model \footnote{https://github.com/Tejas-Deo/Safe-Navigation-Training-Autonomous-Vehicles-using-Deep-Reinforcement-Learning-in-CARLA} to test. 
The algorithm can obtain the nearest undriven point in the drivable path and control the vehicle to drive to that point. In addition, the algorithm can perceive the image data captured by the front camera of the vehicle and process it into a depth map and image segmentation map \cite{xu2019online}. 
After processing by the model, it can output corresponding driving behaviors: accelerate, turn left, turn right, and brake. We plan to use the pre-trained model to test on 50 different difficulty garages generated from garage \autoref{F:origin-m5}. 
We will statistically measure the success rate of testing on each garage and verify and analyze the correlation between the success rate of testing and the difficulty of the garage.

\section{Results}
\label{sec:results}

\subsection{Reinforcement Learning Experiment Results}

\begin{figure*}[!ht] \centering
    \begin{subfigure}[t]{.45\linewidth}
        \centering
        \includegraphics[width=1\textwidth]{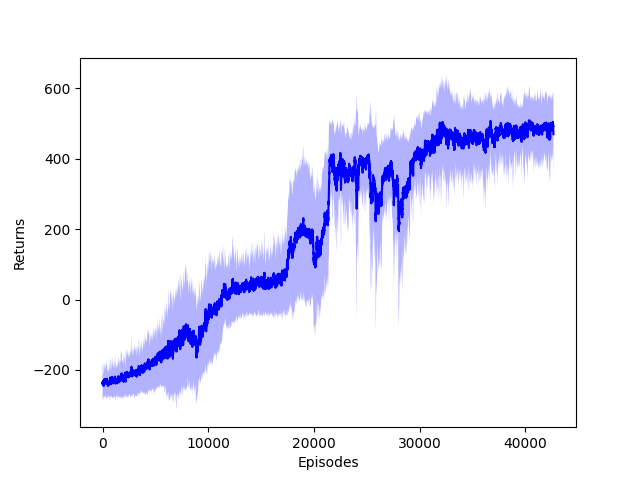}
        \caption{$11\times 7$ Garage}\label{F:plot-m3}
    \end{subfigure}
    \begin{subfigure}[t]{.45\linewidth}
        \centering
        \includegraphics[width=1\textwidth]{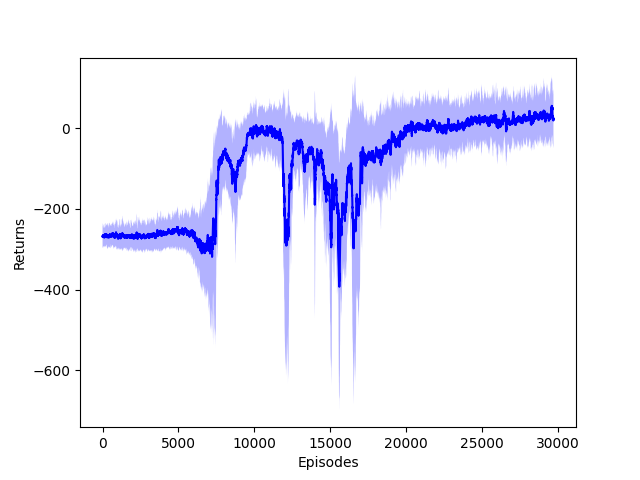}
        \caption{$13\times 13$ Garage without corresponding entrance and exit}\label{F:plot-m4}
    \end{subfigure}
    \begin{subfigure}[t]{.32\linewidth}
        \centering
        \includegraphics[width=1\textwidth,trim={10 10 40 10},clip]{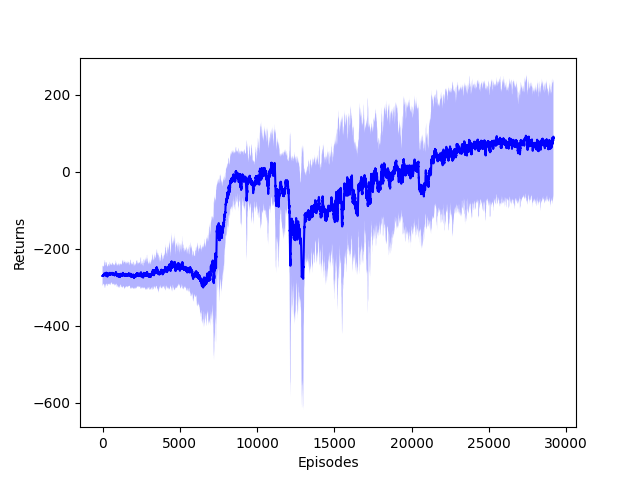}
        \caption{$13\times 13$ Garage with corresponding entrance and exit}\label{F:plot-m5}
    \end{subfigure}
    \begin{subfigure}[t]{.32\linewidth}
        \centering
        \includegraphics[width=1\textwidth,trim={5 10 40 10},clip]{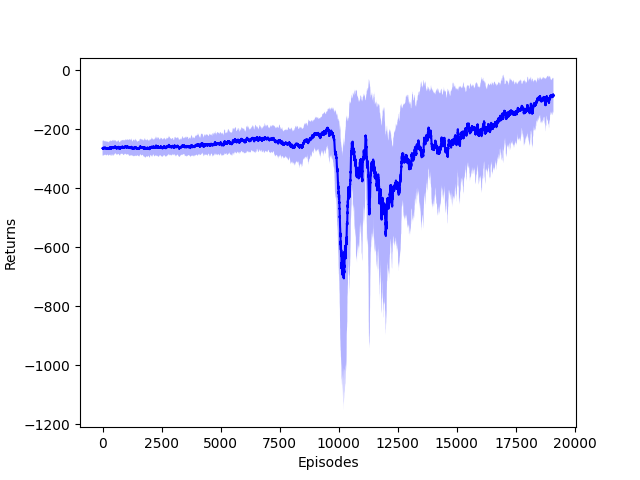}
        \caption{$13\times 13$ S-shaped garage}\label{F:plot-m6}
    \end{subfigure}
    \begin{subfigure}[t]{.32\linewidth}
        \centering
        \includegraphics[width=1\textwidth,trim={10 10 40 10},clip]{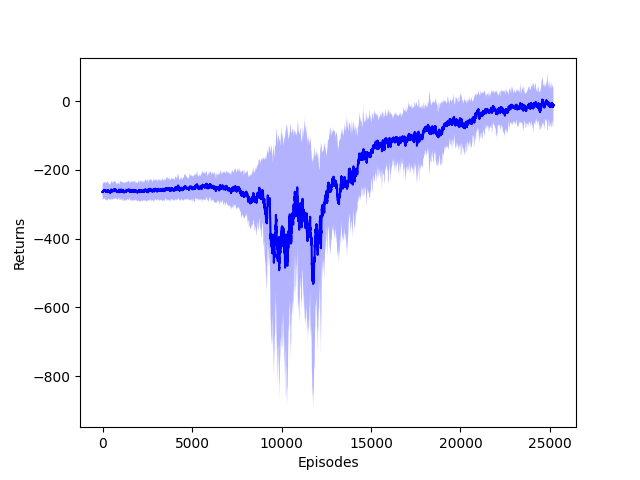}
        \caption{$13\times 13$ U-shaped garage}\label{F:plot-m7}
    \end{subfigure}
    \caption{Convergence curves of reinforcement learning. The x-axis represents the number of episodes, and the y-axis represents the return. For each map, we trained 5 times, took the average of the data for the same episode, and filled the $95\%$ confidence interval. The deep blue line represents the mean, and the light blue area represents the confidence interval.}\label{F:plot-map}
\end{figure*}

The convergence curves of reward for the reinforcement learning experiment on generating encoding are shown in \autoref{F:plot-map}. 
The x-axis represents the number of episodes, and the y-axis represents the return. 
Due to the different sizes and initial states of the garages, the number of episodes for each map is different under the same number of time steps.

It can be observed that the training effects of the $11\times 7$ garage from \autoref{F:plot-m3} and the $13\times 13$ garage with corresponding entrance and exit from \autoref{F:plot-m5} are the best, with the average return values exceeding 0, and the average return of the $11\times 7$ garage exceeds 400, which is relatively high. 
We believe that this is because the search space in the $11\times 7$ garage is relatively small, and the agent easily finds high-reward strategies, while the $13\times 13$ garage requires more training time to ensure stable high rewards.
Among the $13\times 13$ garages, the training effect of the garage with corresponding entrance and exit from \autoref{F:plot-m5} is the best, followed by the garage without corresponding entrance and exit from \autoref{F:plot-m4}, and the worst performance is from the S-shaped and U-shaped garages. 
This is because the structures of the latter two garages are more complex than the former two, and some obstacles block the car from finding the exit while coloring from the entrance.

\begin{figure*}[htb]
    \centering
    \begin{subfigure}[t]{.45\linewidth}
        \centering
        \includegraphics[width=1\textwidth]{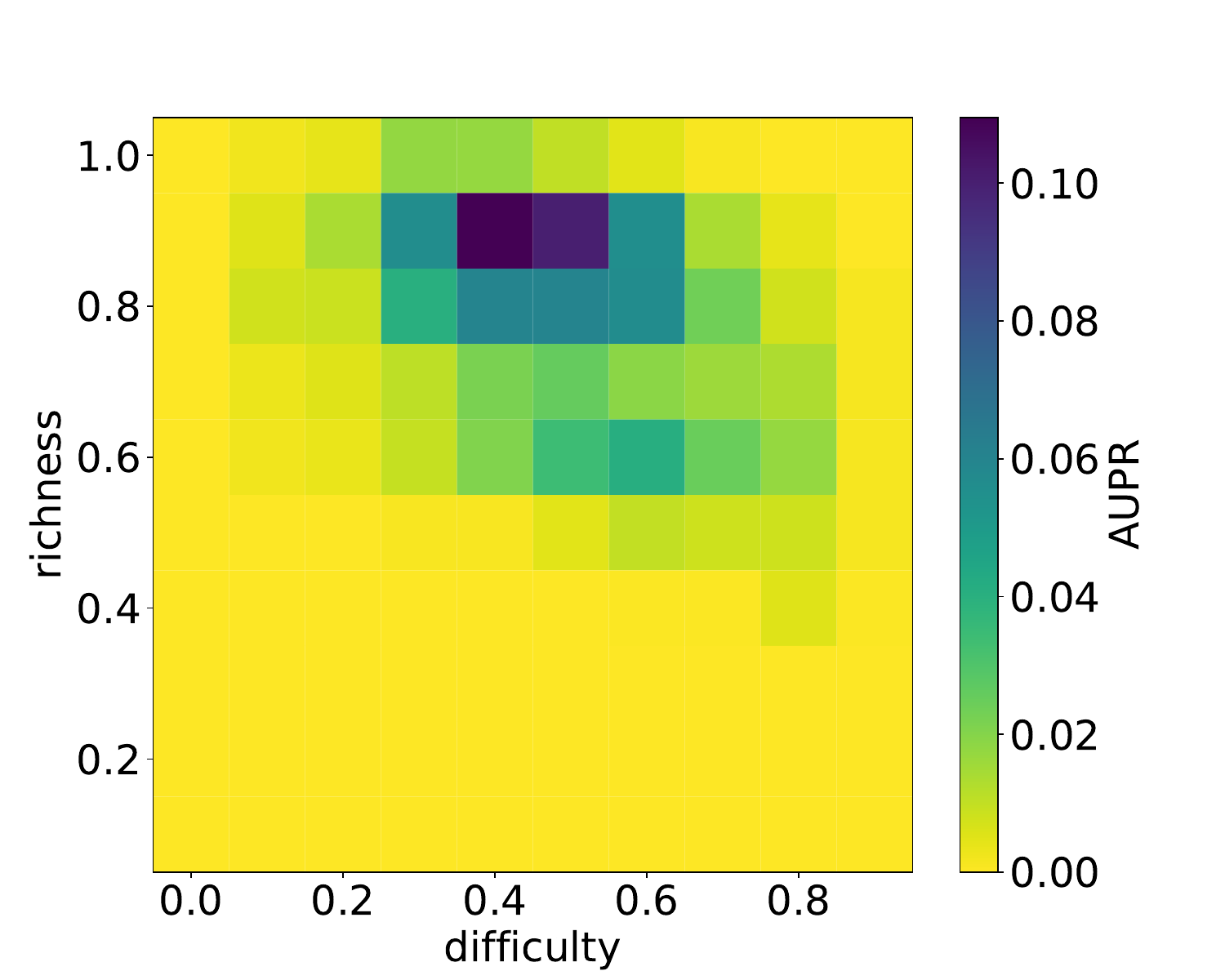}
        \caption{$11\times 7$ Garage}\label{F:heatmap-m3}
    \end{subfigure}
    \begin{subfigure}[t]{.45\linewidth}
        \centering
        \includegraphics[width=1\textwidth]{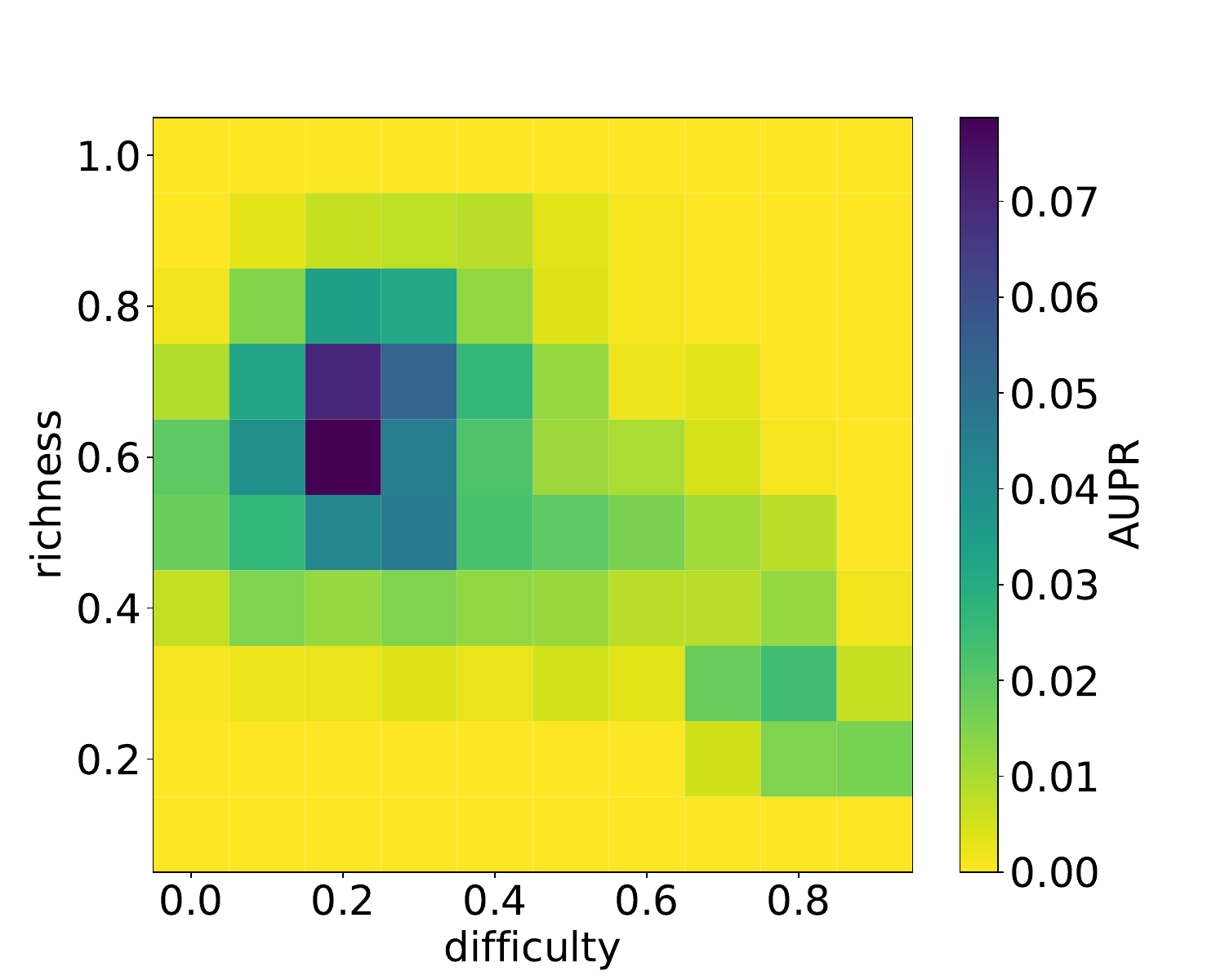}
        \caption{$13\times 13$ Garage without corresponding entrance and exit}\label{F:heatmap-m4}
    \end{subfigure}
    \begin{subfigure}[t]{.32\linewidth}
        \centering
        \includegraphics[width=1\textwidth,trim={20 5 35 10},clip]{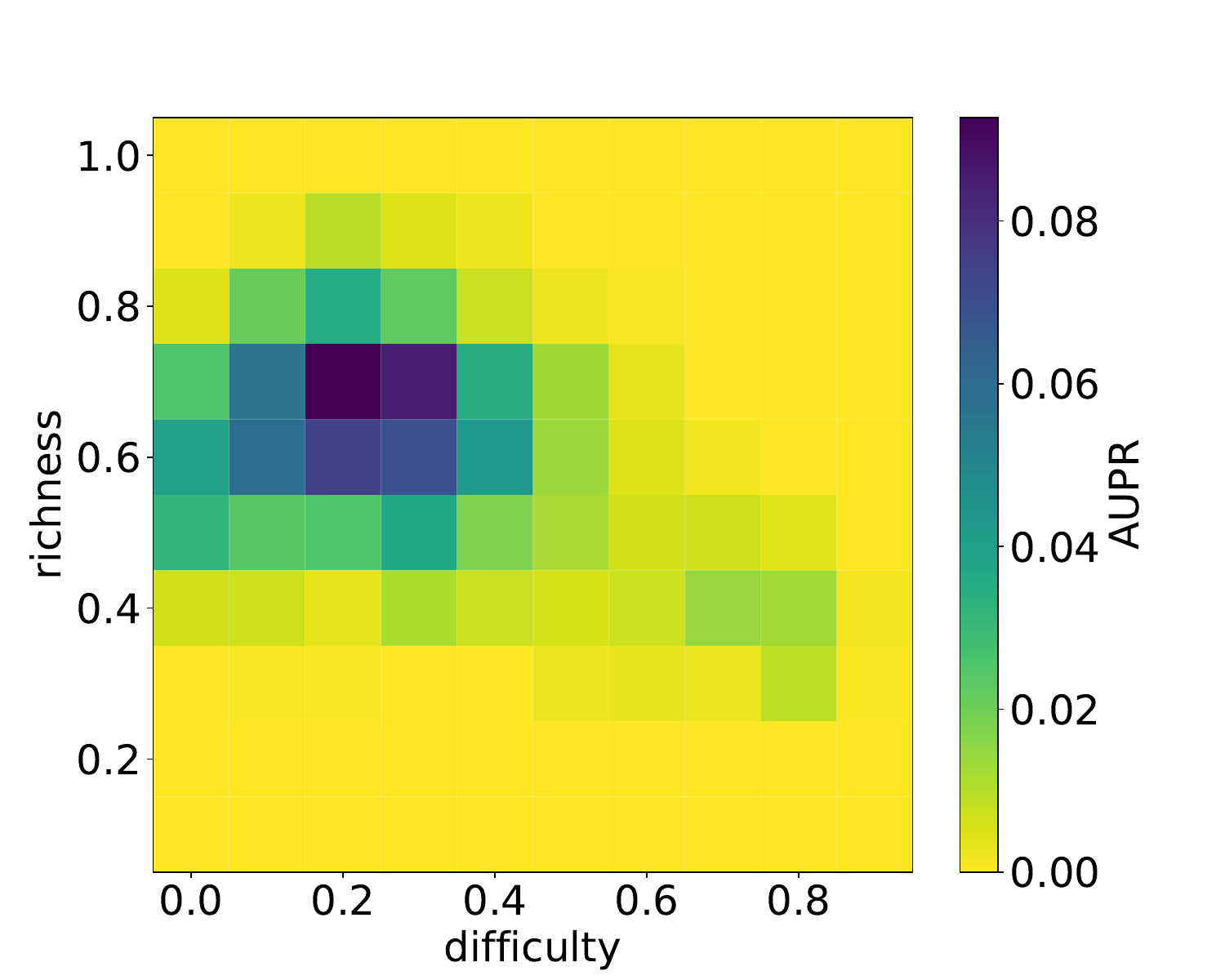}
        \caption{$13\times 13$ Garage with corresponding entrance and exit}\label{F:heatmap-m5}
    \end{subfigure}
    \begin{subfigure}[t]{.32\linewidth}
        \centering
        \includegraphics[width=1\textwidth,trim={20 5 35 10},clip]{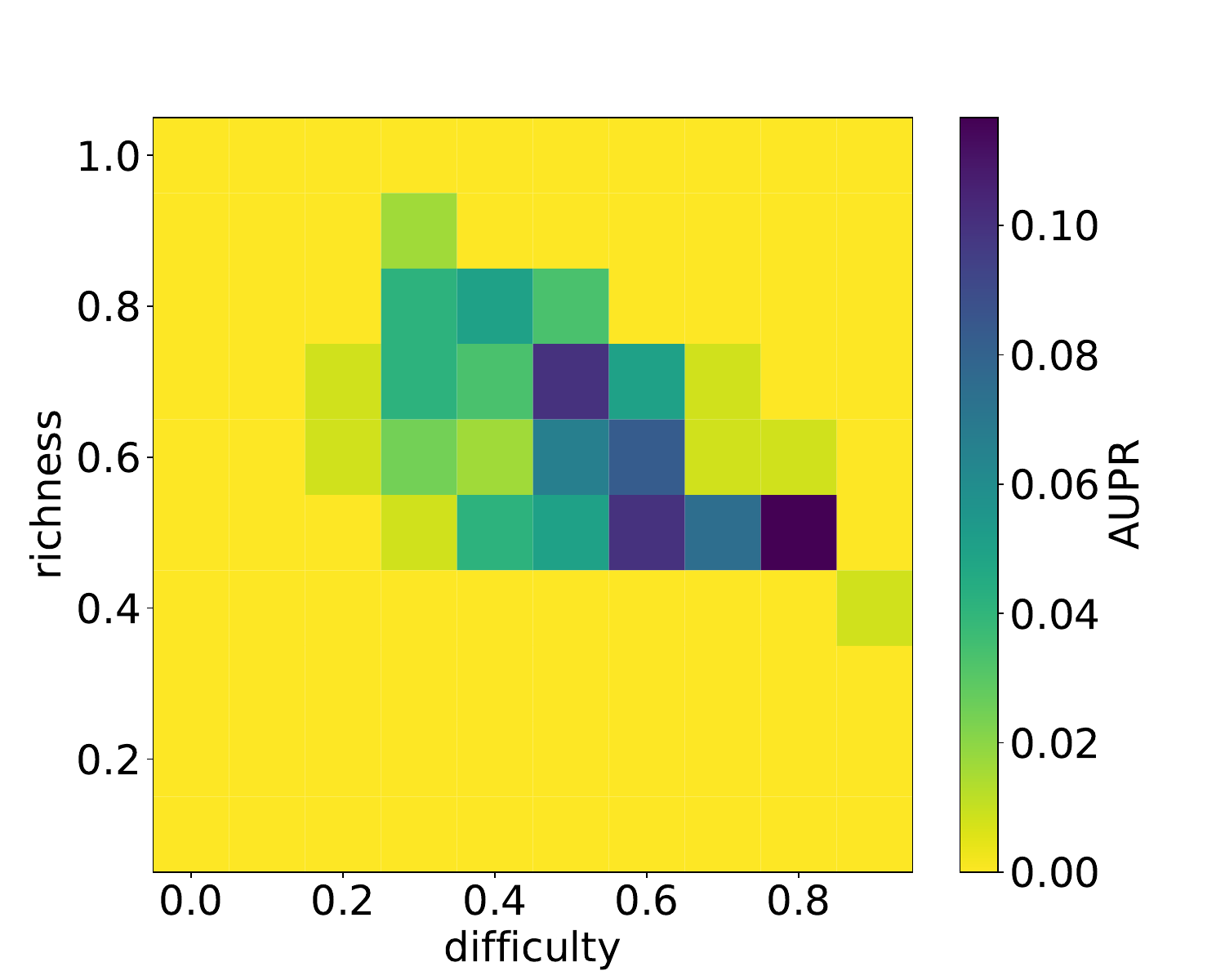}
        \caption{$13\times 13$ S-shaped garage}\label{F:heatmap-m6}
    \end{subfigure}
    \begin{subfigure}[t]{.32\linewidth}
        \centering
        \includegraphics[width=1\textwidth,trim={20 5 35 10},clip]{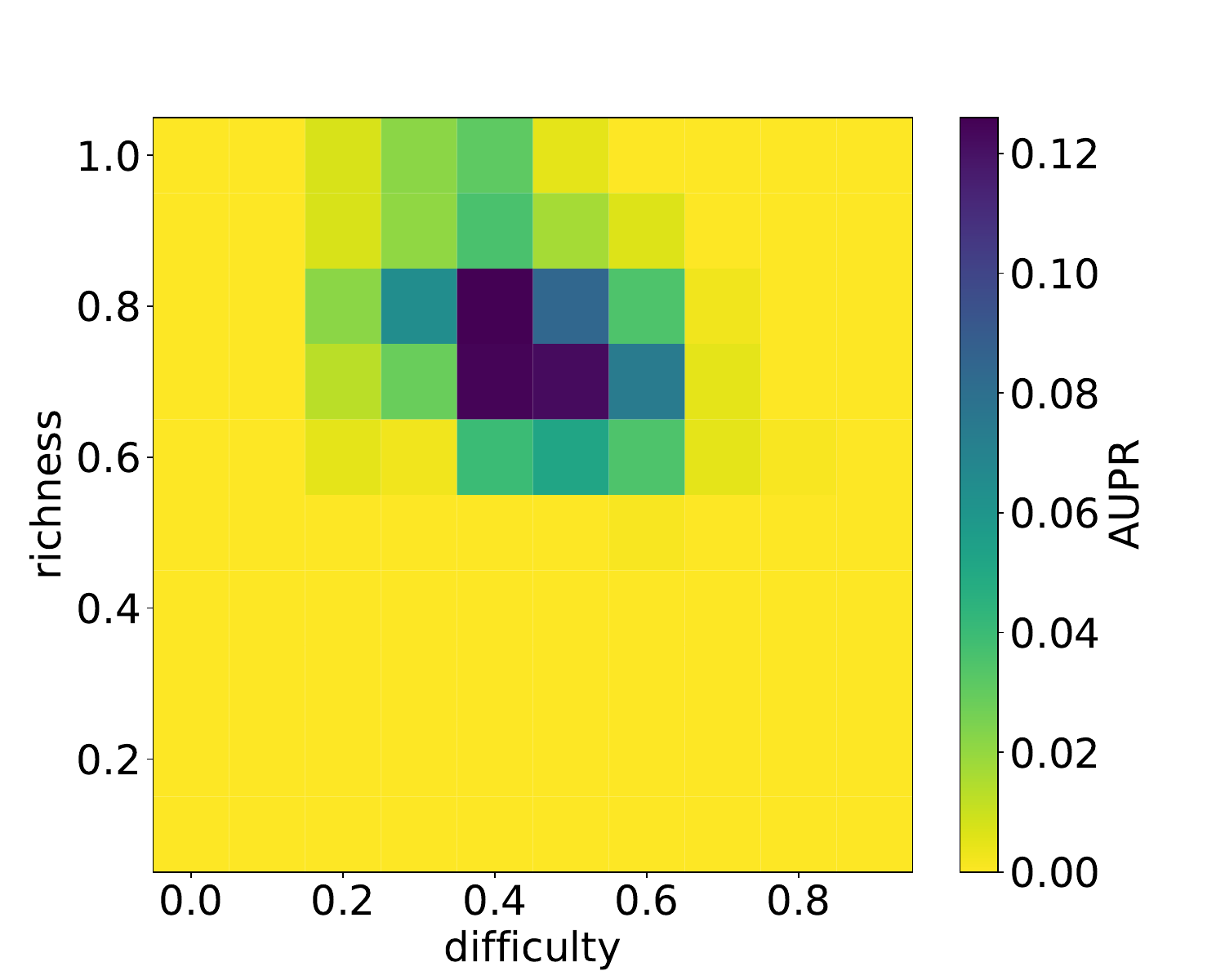}
        \caption{$13\times 13$ U-shaped garage}\label{F:heatmap-m7}
    \end{subfigure}
    \caption{Heatmap of garage distribution. The x-axis represents the difficulty, the y-axis represents the richness, and the spacing between the x-axis and y-axis is 0.1. Darker colors indicate more garages distributed in that block.}\label{F:heatmap}
\end{figure*}

The heatmap of the reinforcement learning experiment on generating encoding is shown in \autoref{F:heatmap}.
It can be observed that the coverage of available garages generated by all maps is concentrated between 0.6 and 0.8, while the difficulty varies among different garages. 
The difficulty ranges of \autoref{F:heatmap-m3}, \autoref{F:heatmap-m4}, and \autoref{F:heatmap-m5} are relatively large, especially \autoref{F:heatmap-m4}, where the difficulty coverage range is the largest, ranging from 0 to 1.
On the other hand, \autoref{F:heatmap-m6} and \autoref{F:heatmap-m7} show a relatively smaller range of difficulty.

\subsection{Autonomous Driving Test Results}

The results of the autonomous driving algorithm test for the garage difficulty verification experiment are shown in Table \autoref{T:data}.
\begin{table*}[!ht]  \centering   \small
    \setlength\tabcolsep{5pt} 
    \renewcommand{\arraystretch}{1.0}
    \caption{Sampling test data for garage set $13\times 13$ with corresponding entrance and exit from \autoref{F:origin-m5}. Index represents the unique identifier. Data in the table are rounded to a maximum of three decimal places.}
    \label{T:data}
    \begin{tabular}{|l|l|l|l|l|l|c|c|c|c|c|c|}  \toprule
        Index & $\delta$ & $n1$ & $\lambda$ & $n2$ & Difficulty & Test Count & Failure Count & Collision & Timeout & Deadlock & Success Rate (\%) \\  \midrule
        48 & 3.8 & 0.1 & 6 & 0 & 0.033 & 50 & 7 & 3 & 0 & 4 & 86 \\
        1315 & 3.25 & 0.375 & 6 & 0 & 0.124 & 50 & 10 & 8 & 0 & 2 & 80 \\
        1347 & 3.6 & 0.2 & 5.4 & 0.15 & 0.167 & 50 & 7 & 7 & 0 & 0 & 86 \\
        1844 & 3.167 & 0.417 & 4.833 & 0.292 & 0.333 & 50 & 5 & 5 & 0 & 0 & 90 \\
        2451 & 3.667 & 0.167 & 5.75 & 0.063 & 0.097 & 50 & 8 & 3 & 0 & 5 & 84 \\
        2452 & 3.2 & 0.4 & 4.833 & 0.292 & 0.328 & 50 & 7 & 5 & 0 & 2 & 86 \\
        2962 & 3.333 & 0.334 & 5 & 0.25 & 0.278 & 50 & 1 & 0 & 0 & 1 & \cellcolor{gray!40}\textbf{98} \\
        3068 & 2.8 & 0.6 & 4.833 & 0.292 & 0.394 & 50 & 2 & 2 & 0 & 0 & 96\\
        3947 & 4 & 0 & 5 & 0.25 & 0.168 & 50 & 4 & 3 & 0 & 1 & 92\\
        4087 & 3 & 0.5 & 4.6 & 0.35 & 0.4 & 50 & 7 & 6 & 0 & 1 & 86 \\
        4591 & 2.667 & 0.667 & 3.714 & 0.572 & 0.603 & 50 & 11 & 11 & 0 & 0 & 78 \\
        4696 & 3.5 & 0.25 & 4.333 & 0.417 & 0.362 & 50 & 1 & 0 & 0 & 1 & 98\\
        4812 & 2.4 & 0.8 & 3.667 & 0.583 & 0.655 & 50 & 11 & 9 & 2 & 0 & 78 \\
        4893 & 2.167 & 0.917 & 3.429 & 0.643 & 0.733 & 50 & 50 & 50 & 0 & 0 & \cellcolor{gray!40}\textbf{0} \\
        4985 & 2.2 & 0.9 & 3.333 & 0.667 & 0.744 & 50 & 16 & 16 & 0 & 0 & 68 \\
        5055 & 2.222 & 0.889 & 2.8 & 0.8 & 0.829 & 50 & 29 & 28 & 1 & 0 & 42\\  \bottomrule
    \end{tabular}
\end{table*}
The index serves as the unique identifier for the dataset generated from the $13\times 13$ entrance and exit corresponding garage. 
It can be observed that the distribution of sampled data in terms of difficulty is not uniform, which is consistent with the heatmap shown in \autoref{F:heatmap-m5} above. 
The majority of failure reasons across almost all garages are concentrated on collisions, with a smaller portion concentrated on deadlock and, finally, timeouts, indicating that the model manages to arrive on time in most cases.
It is noteworthy that more than half of the data have a success rate of over 80\%, indicating that the model can handle most of the sampled garages effectively.

The linear regression fitting results for \autoref{T:data} are as follows. The calculated correlation coefficient between difficulty and success rate is -0.638, suggesting a significant negative linear relationship between the two.
\begin{figure}[!ht] \centering \small
    \includegraphics[width=.47\textwidth,trim={15 15 10 0},clip]{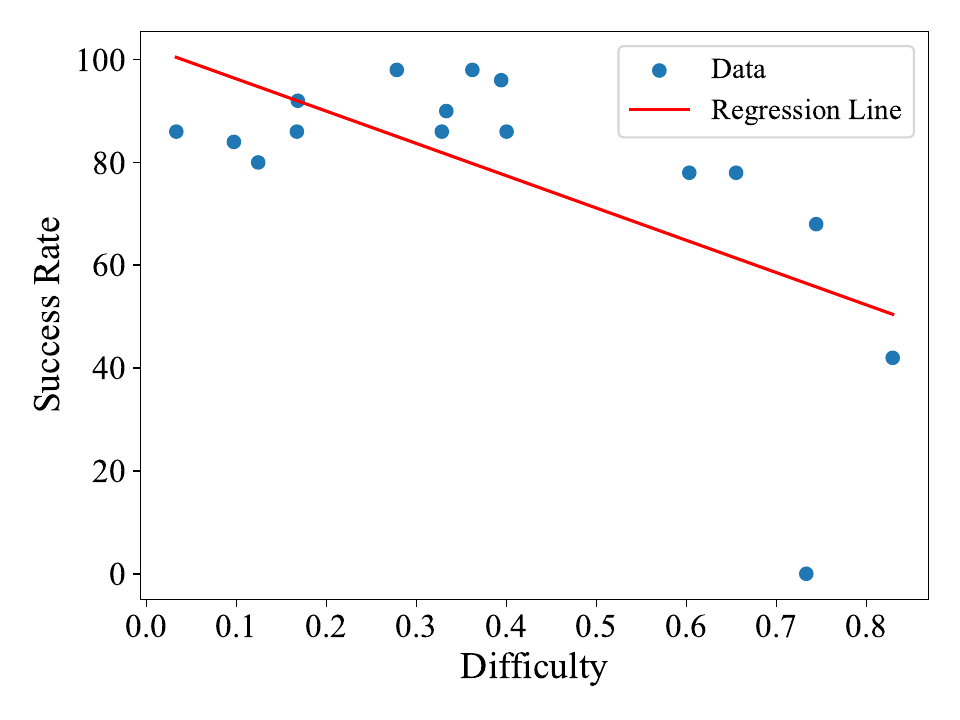}
    \caption{Linear regression fitting graph. The x-axis represents difficulty, and the y-axis represents success rate. The data are from \autoref{T:data}.}\label{F:regression}
\end{figure}

\section{Discussion}
\label{sec:discussion}

We observed that the curves in \autoref{F:plot-m4}, \autoref{F:plot-m5}, \autoref{F:plot-m6}, and \autoref{F:plot-m7} all have a relatively flat segment. 
We believe this is due to the model exploring the penalties associated with road network constraint rules and obstacle collision constraint rules. 
We introduced error indices for these two rules, and when the agent reaches the maximum error index, it receives a significant constant penalty. 
Additionally, we noticed that the curves in \autoref{F:plot-m4}, \autoref{F:plot-m5}, and \autoref{F:plot-m6} have a segment that initially rises and then falls. 
We attribute this to the model gradually exploring the constraints related to moving backward. With each attempt to move backward, the agent receives a small constant penalty. 
However, this penalty has no lower bound, leading the model to fall into extremely negative reward values.

Some maps, such as those shown in the heatmaps in \autoref{F:heatmap-m4} and \autoref{F:heatmap-m5}, contain garages with relatively low coverage. 
This is because the shortest path length connecting the entrance and exit in these maps is small, allowing the agent to search for garages with shorter paths. 
Furthermore, we found that the $11\times 7$ garage, the $13\times 13$ non-corresponding entrance and exit garage, and the $13\times 13$ garage have the largest ranges of difficulty, which aligns well with our expectations.
However, the results for the other two maps struggle to cover more extreme levels of difficulty. 
This is partly related to our reinforcement learning reward function. 
The convergence of reinforcement learning tends to favor garages with medium to low difficulty, and the random walk therein is more likely to produce results similar to or opposite to the convergence results, as can be seen in \autoref{F:heatmap-m6}. 
Additionally, this is related to training time. We believe that reinforcement learning methods can diversify the generation of garages that meet requirements. 
Moreover, as training time increases, the method gradually stabilizes from randomness to exploration, thus exploring more difficulty levels of garages while ensuring their availability.

Based on the linear regression fitting results, we conclude that there is a significant negative linear relationship between difficulty and success rate. 
We believe there are two reasons for this result: 
first, insufficient data; as the number of data samples increases, there are fewer deviating data samples, which better reflect the correlation between the two coefficients. 
Second, the formula and definition of difficulty are inadequate; while difficulty is defined to correspond to the utility rules of reinforcement learning reward functions, it may not intuitively reflect the actual test routes.
Moreover, the difficulty is related to the expected road length and intersection count, but there is still room for adjusting the weights. 
There may be a better weight value, and using this value to calculate difficulty may better reflect its correlation. We believe that, to some extent, higher-difficulty garages are associated with lower model pass rates, which can help identify and correct model issues.

\begin{figure}[htb]
    \centering 
    \begin{subfigure}[t]{.48\linewidth}
        \centering
        \includegraphics[width=1\textwidth]{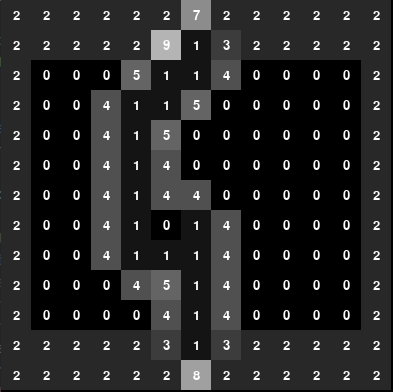}
        \caption{Encoding Matrix}\label{F:i4893}
    \end{subfigure}
    \begin{subfigure}[t]{.48\linewidth}
        \centering
        \includegraphics[width=1\textwidth]{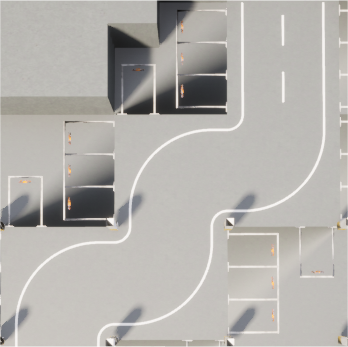}
        \caption{Curvy Road}\label{F:engine-i4893}
    \end{subfigure}
    \caption{The case for garage with index 4893.}\label{F:4893}
\end{figure}

In the simulation tests, some extreme garage scenarios pose challenges. These extreme scenarios can be categorized into overly simple and overly difficult garage scenarios. 
For some highly difficult garages, such as the garage with index 4893 shown in \autoref{F:i4893}, the success rate of the tests can be as low as 0
Observing the structure of this garage, we can see that there is only one path connecting the entrance and exit of the garage. 
Moreover, there is a curvy road near the garage entrance, making it difficult for the model-driven vehicle to navigate through this path.

\begin{figure}[htb]
    \centering
    \includegraphics[width=.45\textwidth]{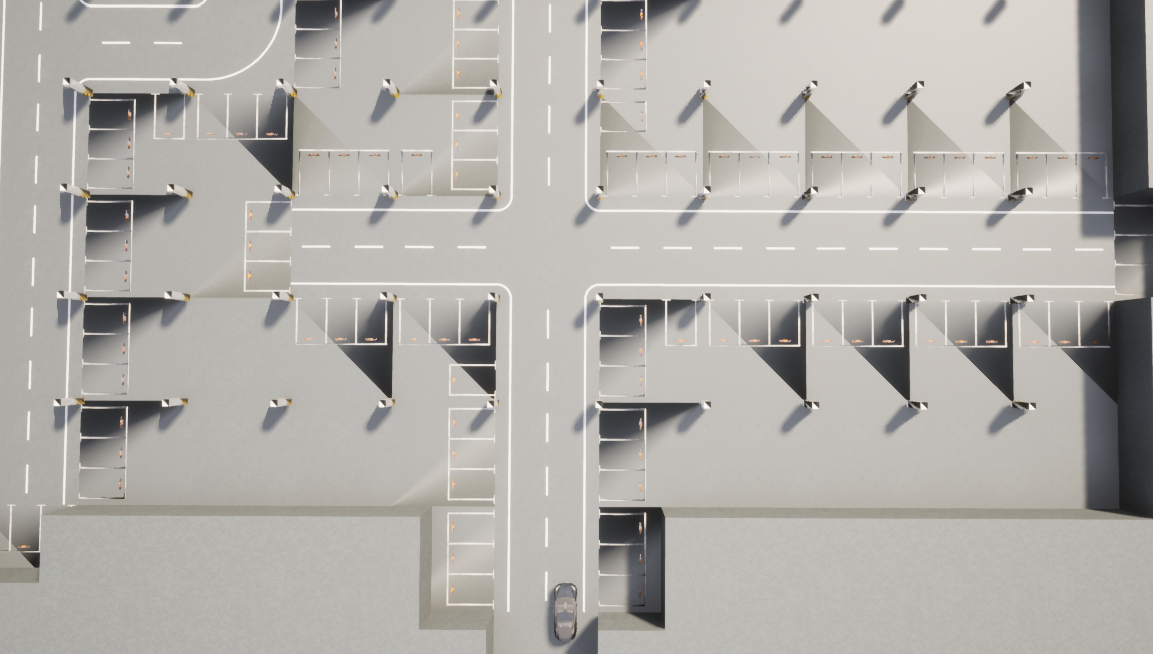}
    \caption{The garage structure with index 2962.}\label{F:2962}
\end{figure}

On the other hand, some garages have a high pass rate. We found that these garages have simpler road structures, as shown in \autoref{F:2962}, with more straight roads and fewer curves, making it easier for the model to navigate. The success rate in such cases can be as high as 98
This result demonstrates the framework's ability to differentiate between algorithm performance on garages with varying difficulty levels, validating the reliability of the difficulty metric. The garage set generated by this framework contributes to improving and optimizing existing AVP algorithms.

\section{Conclusions}
\label{sec:conclusion}

In this paper, we propose a procedural generation framework tailored to the generation of adverse scenarios in static simulation parking environments for autonomous driving. This framework trains a reinforcement learning network based on user-input initial maps and parameters, utilizes deep reinforcement learning methods to generate an encoding matrix dataset, and converts it into FBX and Xodr files.
Ultimately, these files are employed in the Carla simulator to generate three-dimensional parking datasets for training planning algorithms in Autonomous Vehicle Platforms (AVP). 
We establish a mathematical model for this problem, design the environment and reward functions, and construct a comprehensive deep reinforcement learning framework. Two validation metrics, namely coverage and difficulty, are defined for the encoding matrix to evaluate its effectiveness. 
We find that the framework's coverage of the generated parking spaces is predominantly above 0.4, ensuring that the length of paths and the number of parking spaces are at moderate levels. Additionally, the distribution of difficulty for small-sized and obstacle-limited parking spaces is more uniform, indicating the diversity in difficulty levels. 
Experimental testing conducted in the Carla simulator verifies the correlation between the success rate and difficulty, yielding a significant negative linear relationship with a correlation coefficient of -0.648.
Hence, we argue that this framework can generate diverse adverse scenarios in static underground parking environments tailored to the requirements, which can ultimately be utilized to assess the performance of cruise algorithms in AVP and contribute to algorithmic improvement and optimization. 
Through repeated experimentation and profound contemplation, we identify several directions for future improvement:

\begin{enumerate}
    \item Enhance the reward function to improve its correlation with real-world testing scenarios.
    \item Explore the use of multiple deep reinforcement learning methods to obtain better training results.
    \item To optimize parking space generation, the FBX file model should include position-variable static obstacles, dynamic pedestrians, vehicles, etc.. Utilize multiple different planning algorithms to enhance test discrimination.
\end{enumerate}

\bibliographystyle{IEEEtran}
\bibliography{bibliography}

\begin{thebibliography}{10}
\providecommand{\url}[1]{#1}
\csname url@samestyle\endcsname
\providecommand{\newblock}{\relax}
\providecommand{\bibinfo}[2]{#2}
\providecommand{\BIBentrySTDinterwordspacing}{\spaceskip=0pt\relax}
\providecommand{\BIBentryALTinterwordstretchfactor}{4}
\providecommand{\BIBentryALTinterwordspacing}{\spaceskip=\fontdimen2\font plus
\BIBentryALTinterwordstretchfactor\fontdimen3\font minus \fontdimen4\font\relax}
\providecommand{\BIBforeignlanguage}[2]{{%
\expandafter\ifx\csname l@#1\endcsname\relax
\typeout{** WARNING: IEEEtran.bst: No hyphenation pattern has been}%
\typeout{** loaded for the language `#1'. Using the pattern for}%
\typeout{** the default language instead.}%
\else
\language=\csname l@#1\endcsname
\fi
#2}}
\providecommand{\BIBdecl}{\relax}
\BIBdecl

\bibitem{lan2023end}
G.~Lan and Q.~Hao, ``End-to-end planning of autonomous driving in industry and academia: 2022-2023,'' \emph{arXiv e-prints}, pp. arXiv--2401, 2023.

\bibitem{kalra2016driving}
N.~Kalra and S.~M. Paddock, ``Driving to safety: How many miles of driving would it take to demonstrate autonomous vehicle reliability?'' \emph{Transportation Research Part A: Policy and Practice}, vol.~94, pp. 182--193, 2016.

\bibitem{test-begin}
W.~Huang, K.~Wang, Y.~Lv, and F.~Zhu, ``Autonomous vehicles testing methods review,'' in \emph{2016 IEEE 19th International Conference on Intelligent Transportation Systems (ITSC)}, 2016, pp. 163--168.

\bibitem{khastgir2015identifying}
S.~Khastgir, S.~Birrell, G.~Dhadyalla, and P.~Jennings, ``Identifying a gap in existing validation methodologies for intelligent automotive systems: Introducing the 3xd simulator,'' in \emph{2015 IEEE Intelligent Vehicles Symposium (IV)}.\hskip 1em plus 0.5em minus 0.4em\relax IEEE, 2015, pp. 648--653.

\bibitem{dosovitskiy2017carla}
A.~Dosovitskiy, G.~Ros, F.~Codevilla, A.~Lopez, and V.~Koltun, ``Carla: An open urban driving simulator,'' in \emph{Conference on robot learning}.\hskip 1em plus 0.5em minus 0.4em\relax PMLR, 2017, pp. 1--16.

\bibitem{benekohal1988carsim}
R.~F. Benekohal and J.~Treiterer, ``Carsim: Car-following model for simulation of traffic in normal and stop-and-go conditions,'' \emph{Transportation research record}, vol. 1194, pp. 99--111, 1988.

\bibitem{consumed}
A.~Knauss, J.~Schröder, C.~Berger, and H.~Eriksson, ``Paving the roadway for safety of automated vehicles: An empirical study on testing challenges,'' in \emph{2017 IEEE Intelligent Vehicles Symposium (IV)}, 2017, pp. 1873--1880.

\bibitem{blog2021sae}
S.~Blog, ``Sae levels of driving automation refined for clarity and international audience,'' \emph{SAE International}, vol.~3, 2021.

\bibitem{lan2023virtual}
G.~Lan, Q.~Lai, B.~Bai, Z.~Zhao, and Q.~Hao, ``A virtual reality training system for automotive engines assembly and disassembly,'' \emph{IEEE Transactions on Learning Technologies}, 2023.

\bibitem{wiering2012reinforcement}
M.~A. Wiering and M.~Van~Otterlo, ``Reinforcement learning,'' \emph{Adaptation, learning, and optimization}, vol.~12, no.~3, p. 729, 2012.

\bibitem{lan2019simulated}
G.~Lan, J.~Chen, and A.~Eiben, ``Simulated and real-world evolution of predator robots,'' in \emph{2019 IEEE Symposium Series on Computational Intelligence (SSCI)}.\hskip 1em plus 0.5em minus 0.4em\relax IEEE, 2019, pp. 1974--1981.

\bibitem{shaker2016procedural}
N.~Shaker, J.~Togelius, and M.~J. Nelson, ``Procedural content generation in games,'' 2016.

\bibitem{parish2001procedural}
Y.~I. Parish and P.~M{\"u}ller, ``Procedural modeling of cities,'' in \emph{Proceedings of the 28th annual conference on Computer graphics and interactive techniques}, 2001, pp. 301--308.

\bibitem{lindenmayer1968mathematical}
A.~Lindenmayer, ``Mathematical models for cellular interactions in development i. filaments with one-sided inputs,'' \emph{Journal of theoretical biology}, vol.~18, no.~3, pp. 280--299, 1968.

\bibitem{chen2008interactive}
G.~Chen, G.~Esch, P.~Wonka, P.~M{\"u}ller, and E.~Zhang, ``Interactive procedural street modeling,'' in \emph{ACM SIGGRAPH 2008 papers}, 2008, pp. 1--10.

\bibitem{road-network2}
I.~Paranjape, A.~Jawad, Y.~Xu, A.~Song, and J.~Whitehead, ``A modular architecture for procedural generation of towns, intersections and scenarios for testing autonomous vehicles,'' in \emph{2020 IEEE Intelligent Vehicles Symposium (IV)}, 2020, pp. 162--168.

\bibitem{TownSim}
A.~Song and J.~Whitehead, ``Townsim: agent-based city evolution for naturalistic road network generation,'' in \emph{Proceedings of the 14th International Conference on the Foundations of Digital Games}, ser. FDG '19.\hskip 1em plus 0.5em minus 0.4em\relax New York, NY, USA: Association for Computing Machinery, 2019.

\bibitem{lan2022time}
G.~Lan, J.~M. Tomczak, D.~M. Roijers, and A.~Eiben, ``Time efficiency in optimization with a bayesian-evolutionary algorithm,'' \emph{Swarm and Evolutionary Computation}, vol.~69, p. 100970, 2022.

\bibitem{lan2021learning}
G.~Lan, M.~De~Carlo, F.~van Diggelen, J.~M. Tomczak, D.~M. Roijers, and A.~E. Eiben, ``Learning directed locomotion in modular robots with evolvable morphologies,'' \emph{Applied Soft Computing}, vol. 111, p. 107688, 2021.

\bibitem{lan2021learning2}
G.~Lan, M.~van Hooft, M.~De~Carlo, J.~M. Tomczak, and A.~E. Eiben, ``Learning locomotion skills in evolvable robots,'' \emph{Neurocomputing}, vol. 452, pp. 294--306, 2021.

\bibitem{lan2022semantic}
G.~Lan, T.~Liu, X.~Wang, X.~Pan, and Z.~Huang, ``A semantic web technology index,'' \emph{Scientific reports}, vol.~12, no.~1, p. 3672, 2022.

\bibitem{road-network1}
A.~Gambi, M.~Mueller, and G.~Fraser, ``Automatically testing self-driving cars with search-based procedural content generation,'' in \emph{Proceedings of the 28th ACM SIGSOFT International Symposium on Software Testing and Analysis}, ser. ISSTA 2019.\hskip 1em plus 0.5em minus 0.4em\relax New York, NY, USA: Association for Computing Machinery, 2019, p. 318–328.

\bibitem{RoadNetGAN}
T.~Owaki and T.~Machida, ``Roadnetgan: Generating road networks in planar graph representation,'' in \emph{Neural Information Processing}, H.~Yang, K.~Pasupa, A.~C.-S. Leung, J.~T. Kwok, J.~H. Chan, and I.~King, Eds.\hskip 1em plus 0.5em minus 0.4em\relax Cham: Springer International Publishing, 2020, pp. 535--543.

\bibitem{NetGAN}
A.~Bojchevski, O.~Shchur, D.~Z{\"u}gner, and S.~G{\"u}nnemann, ``Netgan: Generating graphs via random walks,'' in \emph{International conference on machine learning}.\hskip 1em plus 0.5em minus 0.4em\relax PMLR, 2018, pp. 610--619.

\bibitem{ferns2003metrics}
N.~F. Ferns, ``Metrics for markov decision processes,'' 2003.

\bibitem{lan2022vision}
G.~Lan, Y.~Wu, F.~Hu, and Q.~Hao, ``Vision-based human pose estimation via deep learning: a survey,'' \emph{IEEE Transactions on Human-Machine Systems}, vol.~53, no.~1, pp. 253--268, 2022.

\bibitem{lan2022class}
G.~Lan, Z.~Gao, L.~Tong, and T.~Liu, ``Class binarization to neuroevolution for multiclass classification,'' \emph{Neural Computing and Applications}, vol.~34, no.~22, pp. 19\,845--19\,862, 2022.

\bibitem{gao2021neat}
Z.~Gao and G.~Lan, ``A neat-based multiclass classification method with class binarization,'' in \emph{Proceedings of the genetic and evolutionary computation conference companion}, 2021, pp. 277--278.

\bibitem{lan2019evolving}
G.~Lan, L.~De~Vries, and S.~Wang, ``Evolving efficient deep neural networks for real-time object recognition,'' in \emph{2019 IEEE Symposium Series on Computational Intelligence (SSCI)}.\hskip 1em plus 0.5em minus 0.4em\relax IEEE, 2019, pp. 2571--2578.

\bibitem{lan2018real}
G.~Lan, J.~Benito-Picazo, D.~M. Roijers, E.~Dom{\'\i}nguez, and A.~Eiben, ``Real-time robot vision on low-performance computing hardware,'' in \emph{2018 15th international conference on control, automation, robotics and vision (ICARCV)}.\hskip 1em plus 0.5em minus 0.4em\relax IEEE, 2018, pp. 1959--1965.

\bibitem{DQN}
V.~Mnih, K.~Kavukcuoglu, D.~Silver, A.~Graves, I.~Antonoglou, D.~Wierstra, and M.~Riedmiller, ``Playing atari with deep reinforcement learning,'' 2013.

\bibitem{stable-baselines3}
A.~Raffin, A.~Hill, A.~Gleave, A.~Kanervisto, M.~Ernestus, and N.~Dormann, ``Stable-baselines3: Reliable reinforcement learning implementations,'' \emph{Journal of Machine Learning Research}, vol.~22, no. 268, pp. 1--8, 2021.

\bibitem{xu2019online}
H.~Xu, G.~Lan, S.~Wu, and Q.~Hao, ``Online intelligent calibration of cameras and lidars for autonomous driving systems,'' in \emph{2019 IEEE Intelligent Transportation Systems Conference (ITSC)}.\hskip 1em plus 0.5em minus 0.4em\relax IEEE, 2019, pp. 3913--3920.

\end{thebibliography}

\end{document}